\documentclass{article}

\usepackage{microtype}
\usepackage{graphicx}
\usepackage{subfigure}
\usepackage{booktabs} 
\usepackage{amsfonts}
\usepackage{amsmath}
\usepackage{mathtools}
\usepackage{subfigure}
\usepackage{enumerate}
\usepackage{algorithm}
\usepackage{algorithmic}
\usepackage{appendix}
\usepackage{hyperref}

\usepackage[accepted]{icml2020}

\icmltitlerunning{Dynamic Knapsack Optimization Towards Efficient Multi-Channel Sequential Advertising}

\begin{document}

\twocolumn[
\icmltitle{Dynamic Knapsack Optimization Towards Efficient\\Multi-Channel Sequential Advertising}



\icmlsetsymbol{equal}{*}

\begin{icmlauthorlist}
\icmlauthor{Xiaotian Hao}{equal,tju}
\icmlauthor{Zhaoqing Peng}{equal,ali}
\icmlauthor{Yi Ma}{equal,tju}
\icmlauthor{Guan Wang}{thu}
\icmlauthor{Junqi Jin}{ali}
\icmlauthor{Jianye Hao}{tju}
\icmlauthor{Shan Chen}{ali}
\icmlauthor{Rongquan Bai}{ali}
\icmlauthor{Mingzhou Xie}{ali}
\icmlauthor{Miao Xu}{ali}
\icmlauthor{Zhenzhe Zheng}{sju}
\icmlauthor{Chuan Yu}{ali}
\icmlauthor{Han Li}{ali}
\icmlauthor{Jian Xu}{ali}
\icmlauthor{Kun Gai}{ali}
\end{icmlauthorlist}

\icmlaffiliation{tju}{College of Intelligence and Computing, Tianjin University, Tianjin, China}
\icmlaffiliation{ali}{Alimama, Alibaba Group, Beijing, China}
\icmlaffiliation{thu}{Department of Automation, Tsinghua University, Beijing, China}
\icmlaffiliation{sju}{Department of Computer Science and Engineering, Shanghai Jiao Tong University, Shanghai, China}

\icmlcorrespondingauthor{Junqi Jin}{junqi.jjq@alibaba-inc.com}
\icmlcorrespondingauthor{Jianye Hao}{jianye.hao@tju.edu.cn}

\icmlkeywords{Recommender Systems, Dynamic Knapsack Problem, Knapsack Problem, Reinforcement Learning, Combinatorial Optimization, Machine Learning, ICML}

\vskip 0.3in
]



\printAffiliationsAndNotice{\icmlEqualContribution} 

\begin{abstract}
In E-commerce, advertising is essential for merchants to reach their target users. The typical objective is to maximize the advertiser's cumulative revenue over a period of time under a budget constraint. In real applications, an advertisement (ad) usually needs to be exposed to the same user multiple times until the user finally contributes revenue (e.g., places an order). However, existing advertising systems mainly focus on the immediate revenue with single ad exposures, ignoring the contribution of each exposure to the final conversion, thus usually falls into suboptimal solutions. In this paper, we formulate the sequential advertising strategy optimization as a dynamic knapsack problem. We propose a theoretically guaranteed bilevel optimization framework, which significantly reduces the solution space of the original optimization space while ensuring the solution quality. To improve the exploration efficiency of reinforcement learning, we also devise an effective action space reduction approach. Extensive offline and online experiments show the superior performance of our approaches over state-of-the-art baselines in terms of cumulative revenue.
\end{abstract}

\section{Introduction}
\label{intro}

In E-commerce, online advertising plays an essential role for merchants to reach their target users, in which Real-time Bidding (RTB) \cite{zhang2014optimal,zhang2016optimal,zhu2017optimized} is an important mechanism. In RTB, each advertiser is allowed to bid for every individual ad impression opportunity. Within a period of time, there are a number of impression opportunities (user requests) arriving sequentially. For each impression, each advertiser offers a bid based on the impression \textbf{{value}} (e.g., revenue) and competes with other bidders in real-time. The advertiser with the highest bid wins the auction and thus display ad and enjoys the impression value. Displaying an ad also associates with a \textbf{{cost}}: in Generalized Second-Price (GSP) Auction \cite{edelman2007internet}, the winner is charged for fees according to the second highest bid. The typical advertising objective for an advertiser is to maximize its cumulative revenue of winning impressions over a time period under a fixed budget constraint.
\begin{figure*}[htbp]
\centering
\includegraphics[height=1.7in, width=6.6in]{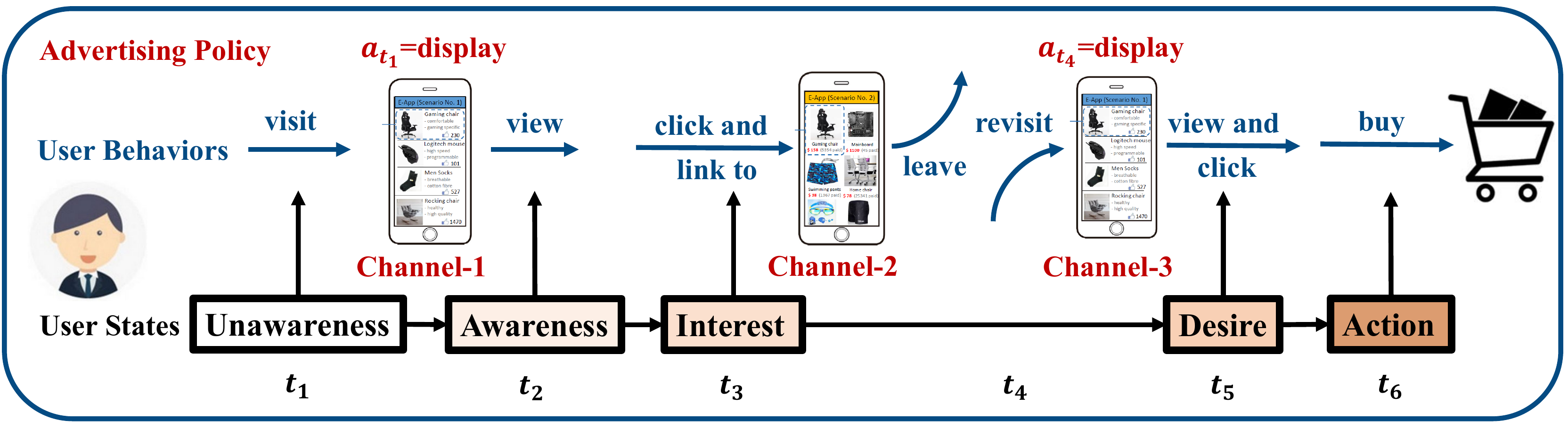}
\caption{An illustration of the sequential multiple interactions (across different channels) between a user and an ad. Each ad exposure has long-term influence on the user's final purchase decision.}
\label{Figure:user-journey}
\end{figure*}

In a digital age, to drive conversion, advertisers can reach and influence users across various channels such as display ad, social ad, paid search ad \cite{ren2018learning}. As illustrated in Figure \ref{Figure:user-journey}, the user's decision to convert (purchase a product) is usually driven by multiple interactions with ads. Each ad exposure would influence the user's preferences and interests, and therefore contributes to the final conversion. However, existing advertising systems \cite{yuan2013real,zhang2014optimal,ren2017bidding,zhu2017optimized,jin2018real,ren2019deep} mainly focus on maximizing the single-step revenue, while ignoring the contribution of previous exposure to the final conversion, and thus usually falls into suboptimal solutions. The reason is that simply optimizing the total immediate revenue cannot guarantee the maximazation of long-term cumulative revenue. Besides, there exist some works \cite{boutilier2016budget,du2017improving,cai2017real,wu2018budget} which optimize the overall revenue under an extra-long (billions) request sequence using a single Constrained Markov Decision Process (CMDP) \cite{altman1999constrained}. However, the optimization of these methods above is myopic as they ignore the mental evolution of each user and long-term advertising effects. The learning is particularly inefficient as well.

Apart from the myopic approaches, there exists some literatures considering the long-term effect of each ad exposure.
Multi-touch attribution (MTA) \cite{ji2017additional,ren2018learning,du2019causally} study the credits assignment to the previous ad displays before conversion. However, these methods only attend to figure out the contribution of each ad exposure, while not providing methods to optimize the strategies. Besides, since all media channels could affect users' conversions,  \citet{li2018efficient,nuara2019dealing} propose multi-channel budget allocation algorithms to help advertisers understand how particular channels contribute to user conversions. They optimize the budget allocation among all channels accordingly to maximize the overall revenue.
However, the granularity of their optimizations is too coarse. They only optimize the budget allocation in the channel level and do not specifically optimize the advertising sequence for each user, which could lead to suboptimal overall performance.

Considering the shortcomings of existing works, we aim at optimizing the budget allocation of an advertiser among all users such that the cumulative revenue of the advertiser could be maximized, by explicitly taking into consideration the long-term influence of ad exposures to individual users. This problem consists of two levels of coupled optimization: bidding strategy learning for each user and budget allocation among users, which we termed as {Dynamic Knapsack Problem}. Different from traditional Knapsack problem, a number of challenges arise: 1) Given the estimated long-term value and cost for each user, the optimization space of the budget allocation grows exponentially in the number of users. Besides, since different advertising policies for each user will lead to different long-term values and costs, the overall optimization space is extremely large. 2) The long-term cumulative value and cost for each user are unknown, which are difficult to make accurate estimations.

To address the above challenges, we propose a novel bilevel optimization framework: Multi-channel Sequential Budget Constrained Bidding (\textbf{MSBCB}), which transforms the original bilevel optimization problem into an equivalent two-level optimization with significantly reduced searching space. The higher-level only needs to optimize over one dimensional variable and the lower-level learns the optimal bidding policy for each user and computes the corresponding optimal budget allocation solution. For the lower-level, we derive an optimal reward function with theoretical guarantee. Besides, we also propose an action space reduction approach to significantly increase the learning efficiency of the lower-level. Finally, extensive offline analyses and online A/B testing conducted on one of the world's largest E-commerce platforms, Taobao, show the superior performance of our algorithm over state-of-the-art baselines.

\section{Formulation: Dynamic Knapsack Problem}
\label{MCBCSB}
Within a time period of $k$ days, we assume that there are $N$ users $\{i\!=\!1,...,N\}$ visiting the E-commerce platform. Each user may interact with the app multiple times and trigger multiple advertising requests. During the sequential interactions between an ad and a user, each ad exposure could influence the user's mind and therefore contributes to the final conversion. Given a fixed ad, for each user $i$, we build a separate Markov Decision Process (MDP) \cite{sutton2018reinforcement} to model its sequential interactions with the same ad. We use $\pi_i$ to denote the advertising policy of the ad towards user $i$, which takes user $i$'s state as input and outputs the auction bid. Details of the MDP will be discussed in Section \ref{user-tareget-rl}. For the fixed ad, we define $V_{G}(i|\pi_i)$ and $V_{C}(i|\pi_i)$ as the expected long-term cumulative value and cost for each user $i$ under policy $\pi_i$. Formally,
\begin{equation}\label{expected-return}
\begin{aligned}
&V_{G}(i|\pi_i) = {\mathbb{E}}\left[G_i|\pi_i\right] = {\mathbb{E}}[\sum_{t=0}^{T_i}v_t|\pi_i] \\
&V_{C}(i|\pi_i) = {\mathbb{E}}\left[C_i|\pi_i\right] = {\mathbb{E}}[\sum_{t=0}^{T_i}c_t|\pi_i]
\end{aligned}
\end{equation}
where $v_t$ and $c_t$ represent the value (i.e., the revenue) and cost obtained from each request $t$ according to policy $\pi_i$, $G_{i}\!=\!\sum_{t=0}^{T_i}v_t$ and $C_{i}\!=\!\sum_{t=0}^{T_i}c_t$ represent the long-term cumulative value and cumulative cost, $T_i$ is the length of the interaction sequence between user $i$ and the current ad.

Given the above definitions, for an advertiser, our target is to maximize its long-term cumulative revenue over $k$ days under a budget constraint $B$, which is formulated as:
\begin{equation}
\label{dynamic knapsack problem}
\begin{aligned}
\max_{\Pi} \max_{\mathcal{X}}\ &\sum_{i=1}^{N} x_{i} ~ V_{G}(i|\pi_i) \\
s.t.~&\sum_{i=1}^{N} x_{i} ~ V_{C}(i|\pi_i) \le B
\end{aligned}
\end{equation}
where $\Pi\!=\!\{\pi_1,..., \pi_N\}$, $\mathcal{X}\!=\!\{x_1,...,x_N\}$, and $x_i\in\{0,1\}$ indicates whether the user $i$ is selected. Since whether displaying an ad to user $i$ does not have any impact on user $j$'s behaviors, $V_{G}(i|\pi_i)$, $V_{C}(i|\pi_i)$ and $\pi_i$ among different users are independent. Thus, given any fixed advertising policy $\Pi\!=\!\{\pi_1,...,\pi_N\}$, $V_{G}(i|\pi_i)$ and $V_{C}(i|\pi_i)$ for each user $i$ are fixed and the inner optimization of Equation (\ref{dynamic knapsack problem}) can be viewed as a classic knapsack problem. The items to be put into the knapsack is the users. However, different advertising policies would lead to different $V_{G}(i|\pi_i)$s and $V_{C}(i|\pi_i)$s for each user, thus here we define Equation (\ref{dynamic knapsack problem}) as a {Dynamic Knapsack Problem} where the value and cost of each item in the knapsack are dynamic. From the perspective of optimization, Formulation (\ref{dynamic knapsack problem}) is a typical bilevel optimization, where the optimization of $\Pi$ is embedded (nested) within the optimization of $\mathcal{X}$.
This bilevel optimization is challenging due to the following reasons:
\begin{enumerate}[(1)]
  \item The optimization space of the joint $\Pi$ is continuous (for the bid space is continuous). The optimization space of $\mathcal{X}$ is discrete, which grows exponentially in the number of users (hundreds of millions). Therefore, the solution space of the combination of $\Pi$ and $\mathcal{X}$ is enormous
      and thus is difficult or even impossible to optimize directly.
  \item The value of $V_{G}(i|\pi_i)$ and $V_{C}(i|\pi_i)$ are unknown and variable, efficient approaches are required to estimate these values online under limited samples.
\end{enumerate}

\section{Methodology: MSBCB Framework}
\label{methodology}
\subsection{Bilevel Decomposition and Proof of Correctness}
\label{UTD}
Based on the above analysis, the bilevel optimization (\ref{dynamic knapsack problem}) is computationally prohibitive and cannot be solved directly. In this paper, we first decompose it into an equivalent two-level sequential optimization process. When taking a fixed policy $\Pi$ as input, we denote the optimal solution of the degraded and static Knapsack Problem as $K=\text{KP}(\Pi)$. Further, the global optimal solution of Problem (\ref{dynamic knapsack problem}) could be defined as:
\begin{equation}
\label{optimized_greedy}
K^* = \max_{\pi_1,\pi_2,...,\pi_N} \text{KP}(\Pi)
\end{equation}
where $\pi_1,...,\pi_N$ are independent variables and $K^*$ is the global optimal solution. To obtain $K^*$, we must firstly specify the form of the function $\text{KP}(\Pi)$.

When taking a fixed policy $\Pi$ as input, computing $\text{KP}(\Pi)$ is a classic static knapsack problem. However, another challenge in online advertising is that the user requests are arriving sequentially in real time and thus real-time decision makings are required. Complicated algorithms (e.g. dynamic programming) are not applicable due to the incompleteness of all users’ values and costs.

On the contrary, the Greedy algorithm could compute a greedy solution without completely knowing the whole set of candidate users beforehand. We will discuss this latter. Besides, the Greedy algorithm can achieve nearly optimal solution in the online advertising \cite{zhang2014optimal,wu2018budget}. As proved by \citet{dantzig1957discrete}, if $\forall i \in 1,...,N$, $V_{C}(i|\pi_i)\! \le\!(1-\lambda)B, 0\!\le\!\lambda\!\le\!1$, i.e., the cumulative cost for each user is much less than the budget, the Greedy algorithm achieves an approximation ratio of $\lambda$, which means the greedy solution is at least $\lambda$ times of the optimal solution $K$. The closer the $\lambda$ gets to 1, the higher the quality of the greedy solution will be. In online advertising, $\lambda$ is usually greater than 99.9\%. Thus, the greedy solution is approximately optimal. We provide the detailed data and proof in Section \ref{proof_greedy} of the Appendix. Therefore, in this paper, we refer to the Greedy algorithm, i.e., $\text{KP}(\Pi)\leftarrow\text{{Greedy}}(\Pi)$.

We define $\text{CPR}_{i}\!=\!\frac{V_{G}(i|\pi_i)}{V_{C}(i|\pi_i)}$ as the Cost-Performance Ratio of each user $i$. The greedy solution is computed by:
\begin{enumerate}[(1)]
\item Sorting all users according to the Cost-Performance Ratio $\text{CPR}_{i}$ in a descending order;
\item Pick users from top to bottom until the cumulative cost violates the budget constraint.
\end{enumerate}
\begin{figure}[ht]
  \centering
  \includegraphics[width=3in, height=1.4in]{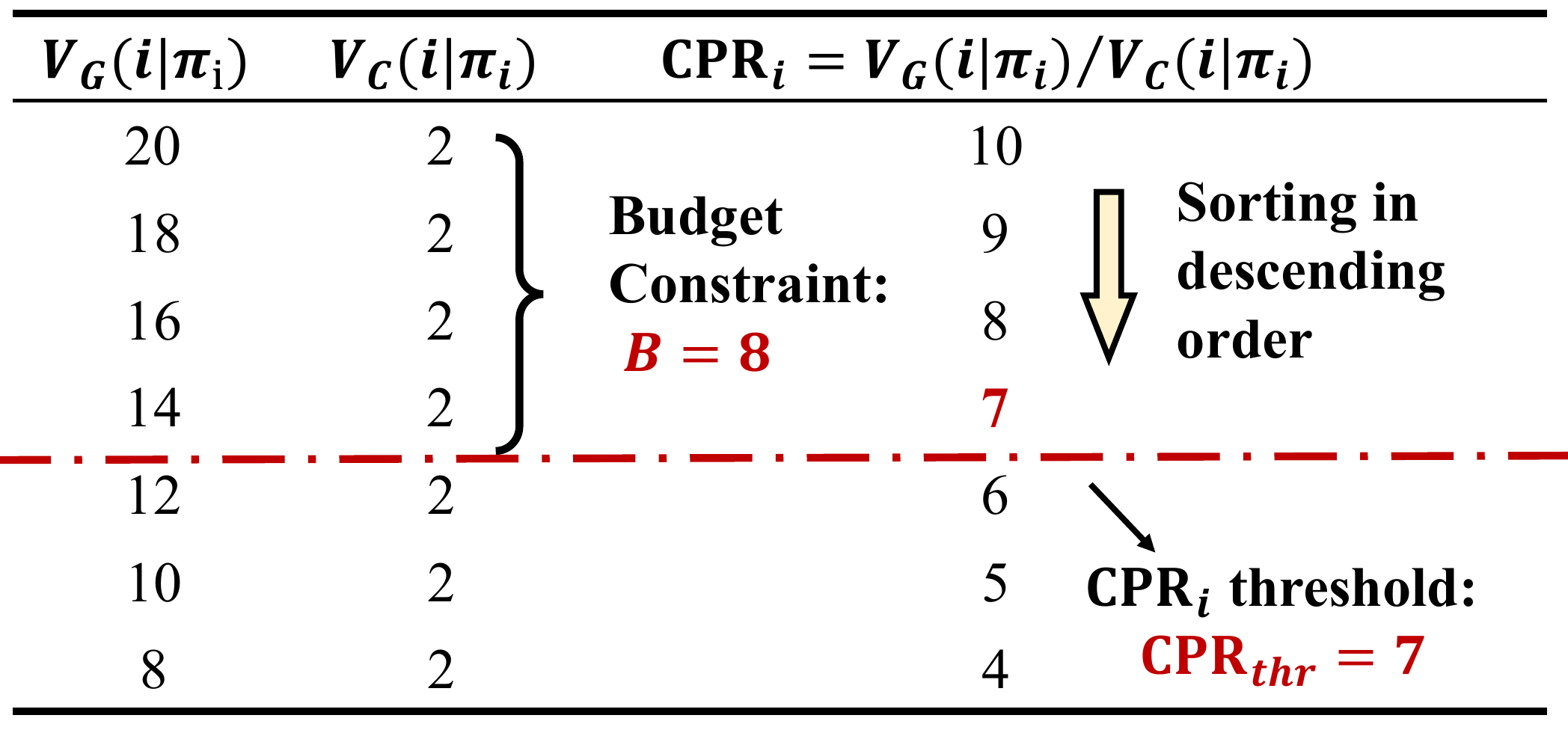}
  \caption{The solution computing process of the Greedy algorithm.}
  \label{greedy-by-lowerbound}
\end{figure}

An illustration is shown in Figure \ref{greedy-by-lowerbound}. In this example, the budget constraint $B=8$. We denote the $\text{CPR}_{i}$ of the last picked user as $\text{CPR}_\text{thr}$, the threshold of the cost-performance ratio. In this example, the $\text{CPR}_\text{thr}=7$. The advantage is that the Greedy algorithm only selects users whose $\text{CPR}_{i}\ge\text{CPR}_\text{thr}$. If we could estimate the $\text{CPR}_\text{thr}$ beforehand, the Greedy algorithm could compute the solution online, without completely knowing the values and costs of all users.

Now that $\text{KP}(\Pi)\leftarrow\text{{Greedy}}(\Pi)$ and the Greedy algorithm prefers users with larger $\text{CPR}_{i}$ (only pick users whose $\text{CPR}_{i}\!\ge\!\text{CPR}_\text{thr}$), according to Equation \ref{optimized_greedy}, to further improve the solution quality, an intuitive way is {to optimize $\pi_i$ for each user $i$ such that each $\text{CPR}_{i}$ could be maximized, i.e., $\pi_i^\prime = \mathop{\text{argmax}}_{\pi_i}\text{CPR}_{i}$.} However, this intuition is incorrect. Maximizing the $\text{CPR}_{i}$ of each user cannot guarantee that the greedy solution $K\!=\!\text{{Greedy}}(\Pi)$ could be maximized. Next, we show that given all users' $\text{CPR}$s are maximized, we can still further improve the solution quality by increasing certain users' allocated budgets and decreasing their $\text{CPR}$s in exchange for greater overall cumulative value. Before we go into the details, we firstly give Lemma 1.
\begin{figure}[ht]
  \centering
  \includegraphics[width=2.8in, height=1.3in]{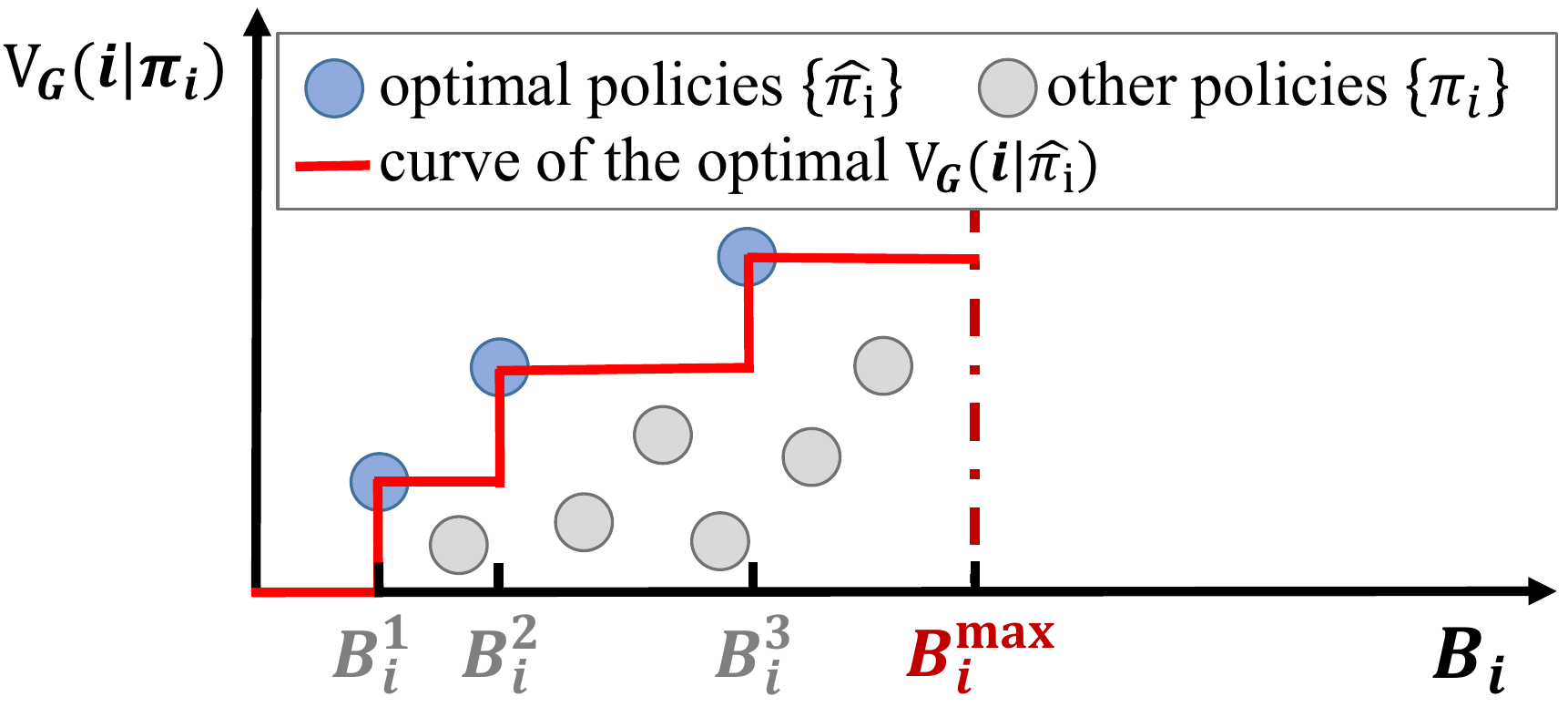}
  \caption{$V_{G}(i|\hat{\pi_i})$ is monotonic with $V_{C}(i|\hat{\pi_i})$.}
  \label{gmv_proportional_cost}
\end{figure}

\textbf{Lemma 1.} \emph{
For each user $i$, the cumulative value $V_{G}(i|\hat{\pi_i})$ increases monotonically with the increase of cost $V_{C}(i|\hat{\pi_i})$ within the range of all possible optimal policies $\{\hat{\pi_i}\}$.
}

{Proof.} We assume that the maximum budget allocated to each user $i$ as $B_i\in[0,B_i^\text{max}]$, where $B_i^\text{max}$ is the maximum cost user $i$ can consume. Then, for each user $i$, within the current budget constraint $B_i$, the optimal advertising policy $\hat{\pi_i}$ must be the one which could maximize the cumulative value, i.e., $\hat{\pi_i}= \mathop{\text{argmax}}_{\pi_i} V_{G}(i|\pi_i),\text{ s.t. } V_{C}(i|\pi_i) \le B_i$. Obviously, as $B_i$ moves from $0$ to $B_i^\text{max}$, we will get a set of optimal policies $\{\hat{\pi_i}\}$, whose cost $V_{C}(i|\hat{\pi_i})$ and value $V_{G}(i|\hat{\pi_i})$ are both increasing. An illustration is shown in Figure \ref{gmv_proportional_cost}. Thus we complete the proof.
\begin{figure}[ht]
  \centering
  \includegraphics[width=3in, height=1.8in]{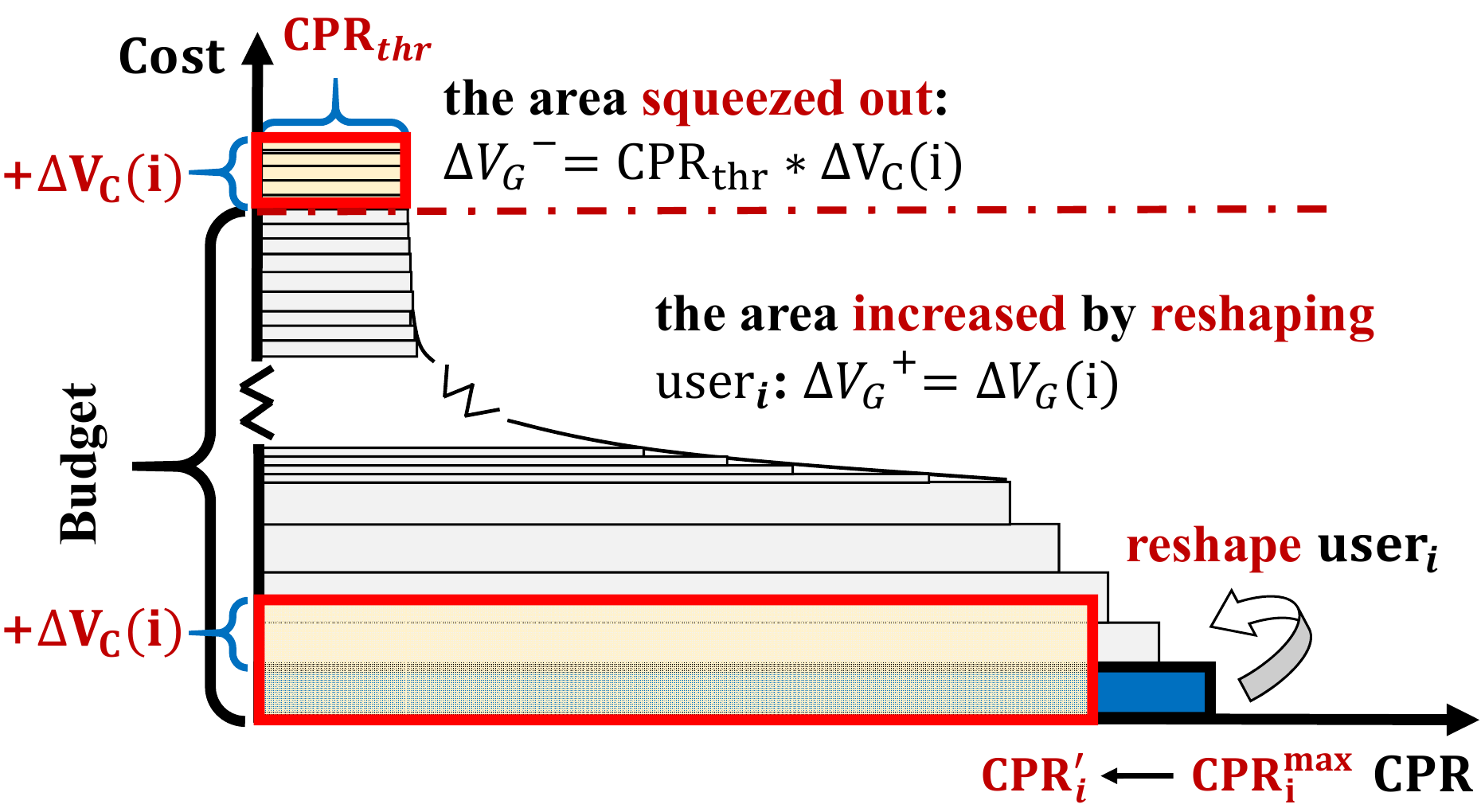}
  \caption{The x-axis denotes each user's $\text{CPR}_i$ and y-axis denotes the cumulative cost of the Greedy algorithm. All users are sorted in descending order and arranged from bottom to top. Each rectangular slice's area (in gray) represents $V_{G}(i|\pi_i)\!=\!\text{CPR}_{i}*V_{C}(i|\pi_i)$, where $\text{CPR}_{i}$ and $V_{C}(i|\pi_i)$ are the width and height. Note that, the height of each rectangular slice is much less than the budget constraint, i.e., $V_{C}(i|\pi_i)\ll B$. The red dashed line marks the position of the budget constraint. The total area of all rectangular slices under the red dashed line constitutes the greedy solution.
}
  \label{Figure:incorrect_of_CPR_max}
\end{figure}

As illustrated in Figure \ref{Figure:incorrect_of_CPR_max}, each user's $\text{CPR}_i$ (the width of each rectangular slice) is maximized initially.
According to Lemma 1, for a user $i$, if we increase $V_{C}(i|\pi_i)$ by $\Delta V_C(i)$, i.e., increase the height of user $i$ by $\Delta V_C(i)$, the corresponding $V_G(i|\pi_i)$ will also increase. We denote this increase in value as $\Delta V_G(i)$. Since there is a budget limit, a small increased height $\Delta V_C(i)$ will squeeze out a small area nearby the $\text{CPR}_\text{thr}$, whose height is also $\Delta V_C(i)$ and width is $\text{CPR}_\text{thr}$ \footnote{Since $\Delta V_C(j) \ll B$, the area squeezed out could be considered as a tiny and smooth change and the width of the last user is approximately equal to $\text{CPR}_\text{thr}$}. We denote the increased area by reshaping user $i$ as $\Delta V_G^+=\Delta V_G(i)$ and the decreased area due to extrusion as $\Delta V_G^-=\text{CPR}_\text{thr}*\Delta V_C(i)$. Overall, if $\Delta V_G^+ > \Delta V_G^-$, the total area will be further increased.
For any user $i$, $\Delta V_G^+ \!>\! \Delta V_G^-$ yields:
\begin{equation}
\label{sufficient_condition}
\begin{aligned}
\Delta V_G(i) &> \text{CPR}_\text{thr}*\Delta V_C(i)
\end{aligned}
\end{equation}
where $\Delta V_G(i)$ and $\Delta V_C(i)$ are caused by the change of $\pi_i$, e.g., from $\pi_i^\prime$ to $\pi_i^{\prime\prime}$.
We denote $\Delta V_G(i)$ as $V_G(i|\pi_i^{\prime\prime})-V_G(i|\pi_i^\prime)$ and $\Delta V_C(i)$ as $V_C(i|\pi_i^{\prime\prime})-V_C(i|\pi_i^\prime)$. We conclude that the greedy solution $K\!=\!\text{{Greedy}}(\Pi^\prime)$ can be further improved if there exists any user $i$ whose current policy $\pi_i^\prime$ can be further improved to $\pi_i^{\prime\prime}$ such that $\Delta V_G(i) > \text{CPR}_\text{thr}*\Delta V_C(i)$. Otherwise, the current solution is optimal. Finally, we provide the definition of the optimal $\pi_i^*$ in Theorem 1.

\textbf{Theorem 1.} \emph{Under the Greedy paradigm ($K\!=\!\text{Greedy}(\Pi)$), for any given $\text{CPR}_\text{thr}$, the optimal advertising policy $\pi_i^*$ for each user $i$ is the one which could maximize $V_G(i|\pi_i)-\text{CPR}_\text{thr}*V_C(i|\pi_i)$. In other words, $\pi_i^*$ is defined as:}
\begin{equation}
\label{optimal-policy-definition}
\begin{aligned}
\pi_i^* = \mathop{\text{argmax}}_{\pi_i} \left[V_{G}(i|\pi_i) - \text{CPR}_\text{thr}*V_{C}(i|\pi_i)\right]
\end{aligned}
\end{equation}
{We denote $\Pi^*=\{\pi_1^*,...,\pi_N^*\}$. The corresponding solution $K_\text{greedy}^*\!=\!\text{Greedy}(\Pi^*)$ is the optimal Greedy solution of the {Dynamic Knapsack Problem} defined in Equation (\ref{dynamic knapsack problem}).}

\textbf{Proof of Theorem 1.} We define $\Pi^*\!=\!\{\pi_1^*,...,\pi_N^*\}$, where $\pi_i^*$ is defined according to Equation (\ref{optimal-policy-definition}), $\forall i \in \{1, ..., N\}$.
We prove Theorem 1 by contradiction. Given the threshold $\text{CPR}_\text{thr}$, we firstly assume that $\text{Greedy}(\Pi^*)$ is not the optimal greedy solution of the {Dynamic Knapsack Problem}, which means we could at least find a user $i$, whose policy $\pi_i^*$ could be further improved to policy $\pi_i^{\prime\prime}$ such that the overall area is increased. This means we could find a better policy $\pi_i^{\prime\prime}$ for user $i$ such that $\Delta V_G(i) > \text{CPR}_\text{thr}*\Delta V_C(i)$ according to Equation (\ref{sufficient_condition}), where $\Delta V_G(i)=V_G(j|\pi_i^{\prime\prime})-V_G(i|\pi_i^*)$ and $\Delta V_C(i)=V_C(i|\pi_i^{\prime\prime})-V_C(i|\pi_i^*)$ ($V_{G}(i|\hat{\pi_i})$ increases monotonically with the increase of $V_{C}(i|\hat{\pi_i})$ according to Lemma 1).
Further, $\Delta V_G(i) > \text{CPR}_\text{thr}*\Delta V_C(i)$ yields:
\begin{equation}
\label{proof-of-the-optimal}
\begin{aligned}
&\left[V_G(i|\pi_i^{\prime\prime}) - \text{CPR}_\text{thr} * V_C(i|\pi_i^{\prime\prime})\right] > \\
 &\left[V_G(i|\pi_i^*) - \text{CPR}_\text{thr} * V_C(i|\pi_i^*)\right]
\end{aligned}
\end{equation}
Equation (\ref{proof-of-the-optimal}) indicates that $$
\pi_i^* \not= \mathop{\text{argmax}}_{\pi_i} \left[V_{G}(i|\pi_i) - \text{CPR}_\text{thr}*V_{C}(i|\pi_i)\right]$$ which contradicts the definition of $\pi_i^*$ in Equation (\ref{optimal-policy-definition}). Thus, the theorem statement is obtained.

\begin{algorithm}[h]
    \caption{MSBCB Framework.}
    \label{Algorithm:two-stage optimization}
    \begin{algorithmic}[1]
        \STATE \textbf{Input:} an initial $\text{CPR}_\text{thr}$;
        \STATE \textbf{Output:} optimal greedy solution of the {Dynamic Knapsack Problem};
        \FOR{each period until convergence}
        \STATE Taking the current estimated $\text{CPR}_\text{thr}$ as input, the agent optimizes the advertising policy $\pi_i$ for each user $i$ according to Section \ref{user-tareget-rl} and acquires the optimal $\Pi^*=\{\pi_1^*,...,\pi_N^*\}$.
        \STATE Based on the current estimated $\text{CPR}_\text{thr}$ and the obtained $\Pi^*$, the agent calculates the greedy solution according to Section \ref{lower level user selection} and collects the actual feedback cost and the predefined budget.
        \STATE Update the estimated $\text{CPR}_\text{thr}$ towards $\text{CPR}_\text{thr}^*$ by minimizing the gap between the actual feedback cost and the budget according to Section \ref{budget_allocation}.
        \ENDFOR
    \end{algorithmic}
\end{algorithm}

We present the overall {MSBCB} framework in Algorithm \ref{Algorithm:two-stage optimization}, which involves a two-level sequential optimization process.
\textbf{(1) Lower-level:} Given any $\text{CPR}_\text{thr}$, we could obtain the optimal advertising policy $\Pi^*$ following Equation \ref{optimal-policy-definition} of Theorem 1, which will be discussed in Section \ref{user-tareget-rl}. Then, based on $\text{CPR}_\text{thr}$ and the optimized $\Pi^*$, we could acquire the Greedy solution by selecting users whose $\text{CPR}_i\ge\text{CPR}_\text{thr}$, which will be detailed in Section \ref{lower level user selection}. \textbf{(2) Higher-level:} However, the current $\text{CPR}_\text{thr}$ might $\neq \text{CPR}_\text{thr}^*$, which means selecting all users whose $\text{CPR}_i\ge\text{CPR}_\text{thr}$ might violate the budget constraint or lead to a substantial budget surplus. Thus, we optimize the current $\text{CPR}_\text{thr}$ towards $\text{CPR}_\text{thr}^*$ in Section \ref{budget_allocation}.
Overall, the optimization space of $\mathcal{X}$ is reduced from $2^N$ to a one-dimensional continuous variable $\text{CPR}_\text{thr}$.
We conclude that Algorithm \ref{Algorithm:two-stage optimization} could iteratively converge to a unique and approximate optimal solution. We present the proof of convergence in Section \ref{proof_convergence} of the Appendix.

\subsection{Lower-level Advertising Policy Optimization with Reinforcement Learning}
\label{user-tareget-rl}
Given a threshold $\text{CPR}_\text{thr}$ as input, we aim to acquire the optimal advertising policy $\pi_i^*$ defined in Equation (\ref{optimal-policy-definition}) of Theorem 1.
Combining the definitions of $V_{G}(i|\pi_i)$ and $V_{C}(i|\pi_i)$ with Equation (\ref{optimal-policy-definition}), we have
\begin{equation}
\label{optimal-policy-definition-derivation}
\begin{aligned}
\pi_i^* =& \mathop{\text{argmax}}_{\pi_i} \left[V_{G}(i|\pi_i) - \text{CPR}_\text{thr}*V_{C}(i|\pi_i)\right]\\
= & \mathop{\text{argmax}}_{\pi_i} \left({\mathbb{E}}[G_i|\pi_i] - \text{CPR}_\text{thr} * {{\mathbb{E}}[C_i|\pi_i]}\right) \\
= & \mathop{\text{argmax}}_{\pi_i} {\mathbb{E}}\left[\left(G_i-\text{CPR}_\text{thr} * C_i\right)|\pi_i\right] \\
= & \mathop{\text{argmax}}_{\pi_i} {\mathbb{E}}[\sum_{t=0}^{T_i}(v_t - \text{CPR}_\text{thr} * c_t)|\pi_i]
\end{aligned}
\end{equation}
Accordingly, we define $r_t=v_t-\text{CPR}_\text{thr}*c_t$, i.e., $\text{value}-\text{CPR}_\text{thr}*\text{cost}$, as the immediate profit acquired at each step $t$. The objective of Equation (\ref{optimal-policy-definition-derivation}) is to obtain the optimal advertising policy $\pi_i^*$ which could maximize the expected long-term cumulative profit. To solve this sequential decision making problem, we formulate it as an MDP and use Reinforcement Learning (RL) \cite{sutton2018reinforcement} techniques to acquire the optimal policy $\pi_i^*$.

We consider an episodic MDP, where an episode starts with the first interaction between a user and an ad, and ends up with a purchase or exceeding the maximum step $T_i$ as:
\begin{itemize}
  \item \textbf{State} $\mathcal{S}$: The state $s_{t}$ should in principle reflect the user request status, ad info, user-ad interaction history info and the RTB environment.
  \item \textbf{Action} $\mathcal{A}$: The action each agent can take in the RTB platform is the bid, which is a real number between 0 and the upper bound $\text{bid}_\text{max}$, i.e., $a_t\in[0, \text{bid}_\text{max}]$.
  \item \textbf{Reward} $\mathcal{R}(\mathcal{S} \times \mathcal{A} \rightarrow \mathbb{R})$: The immediate reward at step $t$ is defined as $r_t=v_t-\text{CPR}_\text{thr}*c_t$.
  \item \textbf{Transition probability} $\mathcal{P}(\mathcal{S} \times \mathcal{A} \times \mathcal{S} \rightarrow [0,1])$: Transition probability is defined as the probability of state transitioning from $s_t$ to $s_{t+1}$ when taking action $a_t$.
  \item \textbf{Discount factor} $\gamma$: The bidding agent aims to maximize the total discounted reward $R_{j}=\sum_{k=t}^{T_i}\gamma\ r_k$ from step $t$ onwards, where $\gamma\in[0, 1]$.
\end{itemize}
For each user $i$, we define the state-action value function $Q(s,a)=\mathbb{E}[R_i|s,a,\pi_i]$ as the expected cumulative reward achieved by following the advertising policy $\pi_i$.
The MDP can be solved using existing Deep Reinforcement Learning (DRL) algorithms such as DQN \cite{mnih2013playing}, DDPG \cite{lillicrap2015continuous} and PPO \cite{schulman2017proximal}. After sufficient training, we would acquire the optimized advertising policies $\Pi^*\!=\!\{\pi_1^*,...\pi_N^*\}$ for all users.
\subsection{Lower-level User Selection by Greedy Algorithm}
\label{lower level user selection}
Taking the current $\text{CPR}_\text{thr}$ and the optimized advertising policies $\Pi^*\!=\!\{\pi_1^*,...\pi_N^*\}$ as inputs, we aim to obtain the greedy solution of the {Dynamic Knapsack Problem}. In reality, we cannot know all users' request sequences and their values and costs beforehand because the user requests are arriving sequentially in real time.
Thus, many complicated methods depending on the completeness of all users' data, e.g., the {dynamic programming approach} \cite{martello1999dynamic}, are not applicable. Even the traditional Greedy algorithm cannot be applied either. Fortunately, the greedy solution could be computed online in an easy way: given the threshold $\text{CPR}_\text{thr}$, the agent only has to select users online whose CPRs are greater than the threshold (an illustration is shown in Figure \ref{greedy-by-lowerbound}). Therefore, we only have to estimate the $\text{CPR}_{i}\!=\!\frac{V_{G}(i|\pi_i)}{V_{C}(i|\pi_i)}$ for each user $i$. To acquire $V_{G}(i|\pi_i)$ and $V_{C}(i|\pi_i)$, besides Q(s,a), we also maintain two other state value functions $V_G(s)$ and $V_C(s)$ according to the Bellman Equation \cite{sutton2018reinforcement}, where $V_G(s)=\mathbb{E}[G_j|s,\pi_j]$ and $V_C(s)=\mathbb{E}[C_j|s,\pi_j]$.

\subsection{Higher-level Optimization by Feedback Control}
\label{budget_allocation}
However, the current estimated threshold $\text{CPR}_\text{thr}$ might have some bias from the optimal $\text{CPR}_\text{thr}^*$. Thus, selecting all users whose $\text{CPR}_i \ge \text{CPR}_\text{thr}$ might violate the budget constraint or lead to a substantial budget surplus. Only when the estimated $\text{CPR}_\text{thr}$ is exactly the same with the optimal $\text{CPR}_\text{thr}^*$, the actual total advertising cost will be equal to the budget. To achieve this, we design a feedback control mechanism, i.e., a PID controller \cite{aastrom1995pid}, to dynamically adjust the $\text{CPR}_\text{thr}$ towards $\text{CPR}_\text{thr}^*$ according to actual feedback of the overall cost. The core formula is:
\begin{equation}
\label{PID}
\resizebox{.905\hsize}{!}{
$\text{CPR}_\text{thr}\ *\!=\left[1\!+\!\alpha_1(\frac{\text{cost}_{t}}{B}\!-\!1)\!+\!\alpha_2(\frac{ \text{cost}_{t\!-\!n:t}}{n*B}\!-\!1)\right]$
}
\end{equation}
where $\text{cost}_{t}$ is the actual feedback cost of the current period, $B$ is the budget, $\text{cost}_{t-n:t}$ and $n\!*\!B$ are the overall cost and the overall budget of the most recent $n$ periods. $\alpha_1$ and $\alpha_2$ are two learning rates. The main idea is when the actual cost exceeds (is less than) the budget, the threshold $\text{CPR}_\text{thr}$ will be increased (decreased) accordingly such that less (more) users will be selected, which will reduce (increase) the cost in turn. The first term $\alpha_1(\frac{\text{cost}_{t}}{B}-1)$ is designed to keep up with the latest changes. The second term $\alpha_2(\frac{ \text{cost}_{t\!-\!n:t}}{n*B}\!-\!1)$ is designed to stabilize learning.

\subsection{Action Space Reduction for RL in Advertising }
\label{action-space-reduction}
However, when applying the RL approaches mentioned in Section \ref{user-tareget-rl} to online advertising, one typical issue is that the sample utilization is inefficient. The main reason is that the action space of the agent is continuous, thus the range of $[0, \text{bid}_\text{max}]$ needs to be fully explored in all states. To resolve this problem, we reduce the magnitude of the continuous action space (i.e., $a_t\in[0, \text{bid}_\text{max}]$) to a binary one (i.e., $\widehat{a_t} \in \{0,1\}$) by making full use of the prior knowledge in advertising, which greatly improves the sample utilization of the RL approaches. Specifically, since different bids $a_t$ can only result in two different outcomes $\widehat{a_t} \in \{0,1\}$,
where $\widehat{a_t}=1$ or 0 indicates whether the ad is displayed to the user, we only have to evaluate the different expected returns resulted by $\widehat{a_t}=1$ or $\widehat{a_t}=0$ for $Q(s,a)$. We denote the greedy action $\widehat{a_t}^*$ based on the current value estimations as:
\begin{equation}
\label{greedy-policy}
\widehat{a_t}^*=\left\{\begin{array}{ll}
\hspace{-1mm}\text{1}&{\text{if } Q(s,\widehat{a_t}=1) > Q(s,\widehat{a_t}=0)} \\
\hspace{-1mm}\text{0}&{\text{otherwise}}
\end{array}\right.
\end{equation}
Then, to obtain an executable bid, for $\widehat{a_t}^*\!=\!0$, we could offer a low enough bid, e.g., $a_t=0$, to make sure that it is impossible to win the auction.
For $\widehat{a_t}^*\!=\!1$, we propose an optimal bid function which could output a bid greater than the second highest bid while not overbidding.

In detail, we maintain two state-action value functions $Q_G(s,\widehat{a_t})\!=\!\mathbb{E}[G_i|s,\widehat{a_t},\pi_i]$ and $Q_C(s,\widehat{a_t})\!=\!\mathbb{E}[C_i|s,\widehat{a_t},\pi_i]$. Since the reward function is defined as $r_t=v_t-\text{CPR}_\text{thr}*c_t$, we have $Q(s,\widehat{a_t})\!=\!Q_G(s,\widehat{a_t})\!-\!\text{CPR}_\text{thr}\!*\!Q_C(s,\widehat{a_t})$.
Then $Q(s,\widehat{a_t}\!=\!1)\!>\!Q(s,\widehat{a_t}\!=\!0)$ yields:
\begin{equation}
\label{condition-1}
\begin{aligned}
&\left[Q_G(s,\widehat{a_t}=1)-\text{CPR}_\text{thr}*Q_C(s,\widehat{a_t}=1)\right] > \\
&\left[Q_G(s,\widehat{a_t}=0)-\text{CPR}_\text{thr}*Q_C(s,\widehat{a_t}=0)\right]
\end{aligned}
\end{equation}

If $\widehat{a_t}=0$, the expected immediate cost is 0 (since the ad is not exposed). If $\widehat{a_t}=1$, we denote the expected immediate cost as $\mathbb{E}[c_t|\widehat{a_t}=1]$, whose value depends on the pricing model. In online advertising, typical pricing models includes CPM (Cost Per Mille, the advertiser bid for impressions and is charged based on impressions), CPC (Cost Per Click, the advertiser bid for clicks and is charged based on clicks) and CPS (Cost Per Sales, the advertiser bid for conversions and is charged based on conversions). If CPM is used, $\mathbb{E}[c_t|\widehat{a_t}=1]=\textbf{bid}_{t}^{\textbf{2nd}}$, where $\textbf{bid}_{t}^{\textbf{2nd}}$ denotes the second highest bid in the auction. If CPC is used, $\mathbb{E}[c_t|\widehat{a_t}=1]=\textbf{bid}_{t}^{\textbf{2nd}} * \text{pCTR}$, where pCTR represents the predicted Click-Through Rate. If CPS is used, $\mathbb{E}[c_t|\widehat{a_t}=1]=\textbf{bid}_{t}^{\textbf{2nd}} * \text{pCTR} * \text{pCVR}$, where pCVR represents the predicted Conversion Rate. For ease of presentation, we take CPM for an example. Under CPM,
\begin{equation}
\label{condition-2}
\begin{aligned}
Q_C(s,\widehat{a_t}=1)=\mathbb{E}[c_t + \sum_{k=t+1}^{T_i}c_k|s,\widehat{a_t}=1,\pi_i]& \\
=\textbf{bid}_{t}^{\textbf{2nd}} + \mathbb{E}[\sum_{k=t+1}^{T_i}c_k|s,\widehat{a_t}=1,\pi_i]& \\
Q_C(s,\widehat{a_t}=0)
=\textbf{0} + \mathbb{E}[\sum_{k=t+1}^{T_i}c_k|s,\widehat{a_t}=0,\pi_i]&
\end{aligned}
\end{equation}
Notice that the second highest bid $\textbf{bid}_{t}^{\textbf{2nd}}$ is unknown until the current auction is finished. Substituting Equation (\ref{condition-2}) into Equation (\ref{condition-1}), we acquire
\begin{equation}
\label{optimal-bid-action}
\begin{aligned}
\textbf{bid}_{t}^{\textbf{2nd}} < &\left[\left(\frac{Q_G(s,\widehat{a_t}=1)}{\text{CPR}_\text{thr}}-Q_C^{\text{next}}(s,\widehat{a_t}=1)\right)\right.\\
-&\left.\left(\frac{Q_G(s,\widehat{a_t}=0)}{\text{CPR}_\text{thr}}-Q_C^{\text{next}}(s,\widehat{a_t}=0)\right)\right]
\end{aligned}
\end{equation}
where $Q_C^{\text{next}}(s,\widehat{a_t})=\mathbb{E}[\sum_{k=t+1}^{T_j}c_k|s,\widehat{a_t},\pi_i]$. We denote the term on the right of the '$<$' in Equation (\ref{optimal-bid-action}) as $\textbf{b}_t^*$. And we conclude that the bidding agent can always set the bid price $a_t=\textbf{b}_t^*$ during the online bidding phase, which is the optimal action without any loss of accuracy. Refer to Section \ref{proof_optimal_bid} of the Appendix for proof. For CPC or CPS, the optimal bid formula $\textbf{b}_t^*$ can be easily acquired by substituting the corresponding $\mathbb{E}[c_t|\widehat{a_t}=1]$ into Equation \ref{condition-2}. Here, we reaffirm that our action space reduction technique is a generalized design and is applicable to different pricing models.

\section{Empirical Evaluation: Simulations}
\label{Simulations}
We start with designing simulation experiments to shed light on the contributions of the proposed framework {MSBCB} under more controlled settings. Similar to the simulation settings of \cite{ie2019reinforcement}, we assume there are a set of users $\{i=1,...,N\}$, a set of ads $\mathcal{D}$ and a set of commodity categories $\mathcal{T}$. Each ad $d \in\mathcal{D}$ has an associated category. Each user $i$ has various degrees of interests in commodity categories, which is influenced by the displayed ad. When user $i$ consumes ad $d$, his interest in category $T(d)$ is nudged stochastically, biased slightly towards increasing his interest, but allows some chance of decreasing his interest. We set $N=10000$, $|\mathcal{D}|=2000$ and $|\mathcal{T}|=20$ in the following experiments. Detailed settings of the simulation environment can be found in Section \ref{offline_env} of the Appendix.

\subsection{Baselines}
We compare our {MCBCB} with following baseline strategies:
\begin{itemize}
  \item {Myopic Approaches:} (1) {Manual Bid} is a strategy that the agent continuously bids at the same price initialized by the advertiser. (2) {Contextual Bandit} \cite{zhang2014optimal} aims at maximizing the accumulated short-term value of each request based on the Greedy framework.
  \item {Greedy with maximized CPR:} This approach is similar to our method under the Greedy framework except that each $\pi_i$ is optimized by maximizing the long-term CPR. In the offline simulation, we enumerate all policies for each user and select the one which could maximize its CPR. This approach is named as {Greedy+maxCPR}.
  \item {Greedy with state-of-the-art RL approaches:} These baselines, i.e., {Greedy+DQN}, {Greedy+DDPG} and {Greedy+PPO}, utilize the same reward function with our {MSBCB} to optimize the lower-level optimization of $\Pi$. The difference is that our {MSBCB} leverages the action space reduction technique. For DQN and PPO, we discretize the bid action space $[0,\text{bid}_\text{max}]$ evenly into 11 real numbers as the valid actions.
  \item {Undecomposed Optimization:} These baselines are RL approaches (DQN,DDPG and PPO) based on the Constrained Markov Decision Process (CMDP). They are named as {Constrained+DQN}, {Constrained+DDPG}, {Constrained+PPO} respectively. We follow the CMDP design and settings in \cite{wu2018budget}.
  \item {Offline Optimal:} The optimal solution of the {Dynamic Knapsack Problem} can be computed by {dynamic programming} in offline simulation because we could enumerate all possible policies to get the corresponding long-term values and costs for each user. Note that since users' request sequences are unknown beforehand and there is only one chance for the ad to bid for each request in the online advertising systems, the optimal solution can only be obtained in offline simulation.
\end{itemize}

\subsection{Experimental Results}
\label{offline_results}
We conduct extensive analysis of our {MSBCB} in the following 5 aspects. All approaches aim to maximize the advertiser's cumulative revenue under a fixed budget constraint. All experimental results are averaged over 10 runs.
The hyperparameters for each algorithm are set to the best we found after grid-search optimization.

\begin{figure}[ht]
  \centering
  \includegraphics[width=2.5in,height=1.8in]{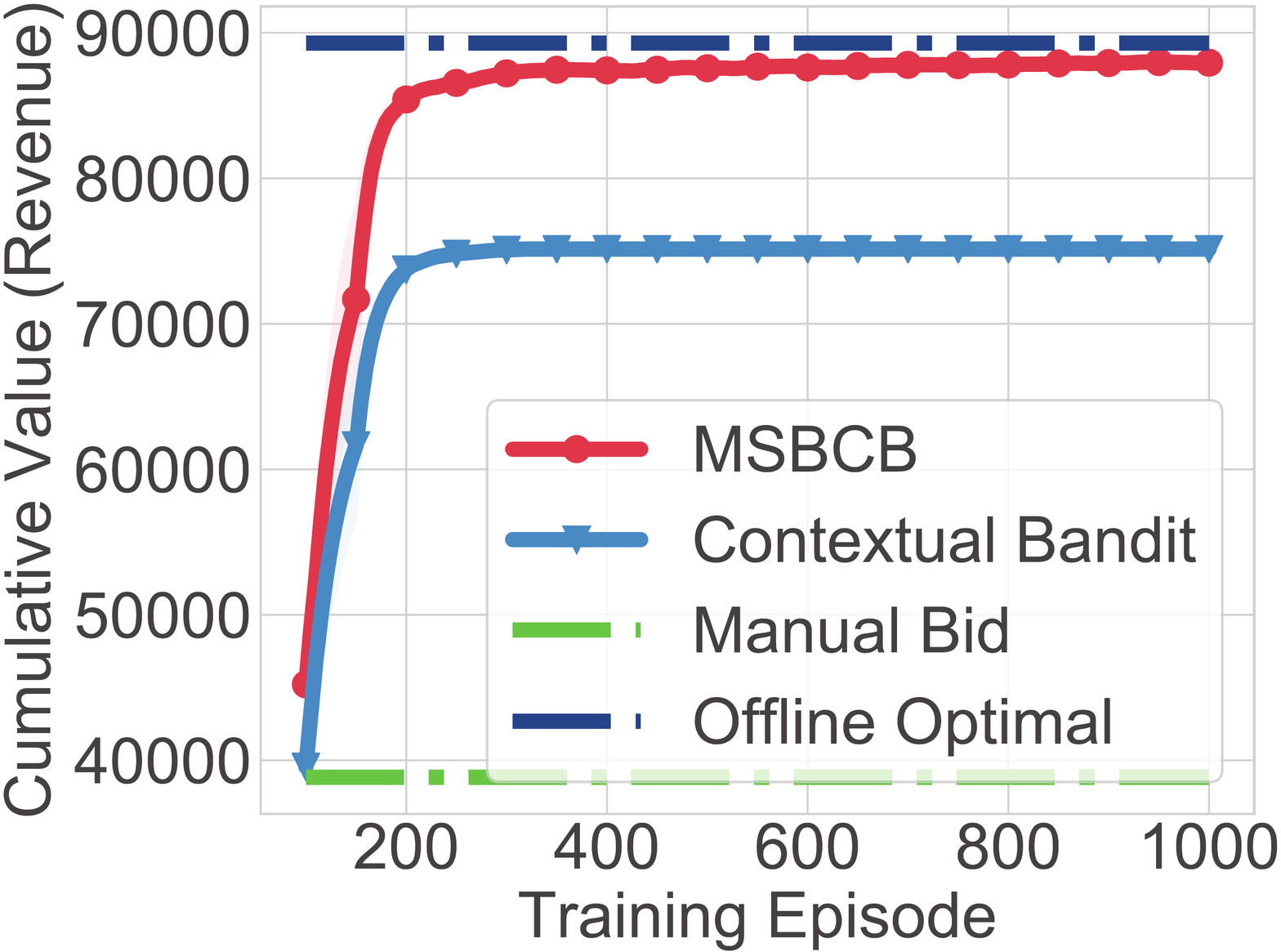}
  \caption{Values comparisons (learning curves) of the myopic approaches with non-myopic approaches and the offline optimal.}
  \label{exp_sub_figure_1}
\end{figure}

\textbf{Myopic vs Non-myopic.}
To show the benefits of upgrading the myopic advertising system into a farsighted one, we compare the cumulative revenue achieved by our {MSBCB} with two other myopic baselines. The learning curves and results are shown in Figure \ref{exp_sub_figure_1} and Table \ref{tab:all_method_results}. We see that {MSBCB} outperforms the {Manual Bid} and the {Contextual Bandit} by a large margin, which indicates that taking account of the long-term effect of each ad exposure could significantly improve the cumulative advertising results.

\textbf{MSBCB vs the Offline Optimal.} In Figure \ref{exp_sub_figure_1}, we also compare our {MSBCB} with the {Offline Optimal}, which is computed by a modified dynamic programming algorithm. We see that as the training continues, our {MSBCB} gradually achieves an approximately optimal solution. Detailed results are summarized in Table \ref{tab:all_method_results}. Our {MSBCB} empirically achieves an approximation ratio of 98.53\%($\pm$0.36\%).

\begin{figure}[ht]
  \centering
  \includegraphics[width=2.5in,height=1.8in]{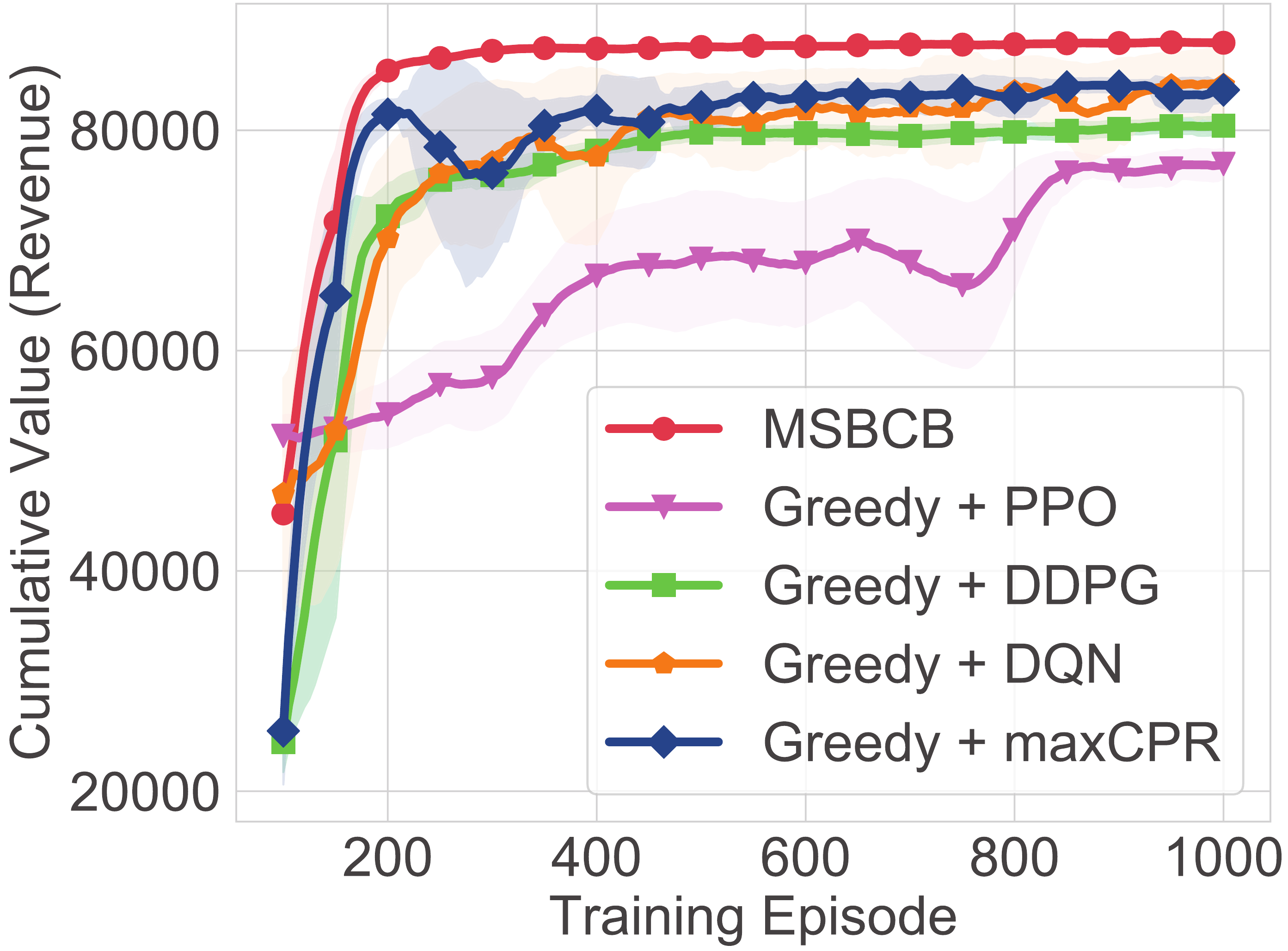}
  \caption{Value comparisons of {MSBCB} with the {Greedy with maximized CPR} and the {Greedy with state-of-the-art RL}.}
  \label{exp_sub_figure_2}
\end{figure}

\textbf{MSBCB vs Greedy with maximized CPR.}
As discussed in Section \ref{UTD}, under the Greedy framework, maximizing each user's $\text{CPR}_{i}$ cannot guarantee that the greedy solution of the {Dynamic Knapsack Problem} (\ref{dynamic knapsack problem}) could be maximized. The optimal advertising policy $\pi_i$ for each user is given by Theorem 1. To experimentally verify the correctness of Theorem 1, we compare the cumulative revenue achieved by {MSBCB} and the {Greedy with maximized CPR}. As shown in Figure \ref{exp_sub_figure_2} and Table \ref{tab:all_method_results}, {MSBCB} outperforms {Greedy with maximized CPR} and achieves a $+5.11\%$ improvement.

\textbf{MSBCB vs Greedy with state-of-the-art RL approaches.}
Besides, to show the effectiveness of the action-space reduction proposed in Section \ref{action-space-reduction}, we compare {MSBCB} with the state-of-the-art DRL approaches under the Greedy framework. As shown in Figure \ref{exp_sub_figure_2} and Table \ref{tab:all_method_results}, MSBCB outperforms {Greedy+DQN}, {Greedy+DDPG} and {Greedy+PPO} both in the cumulative revenue and the convergence speed, which shows that the action space reduction effectively improves the sample efficiency of RL approaches.
\begin{figure}[ht]
  \centering
  \includegraphics[width=2.5in,height=1.8in]{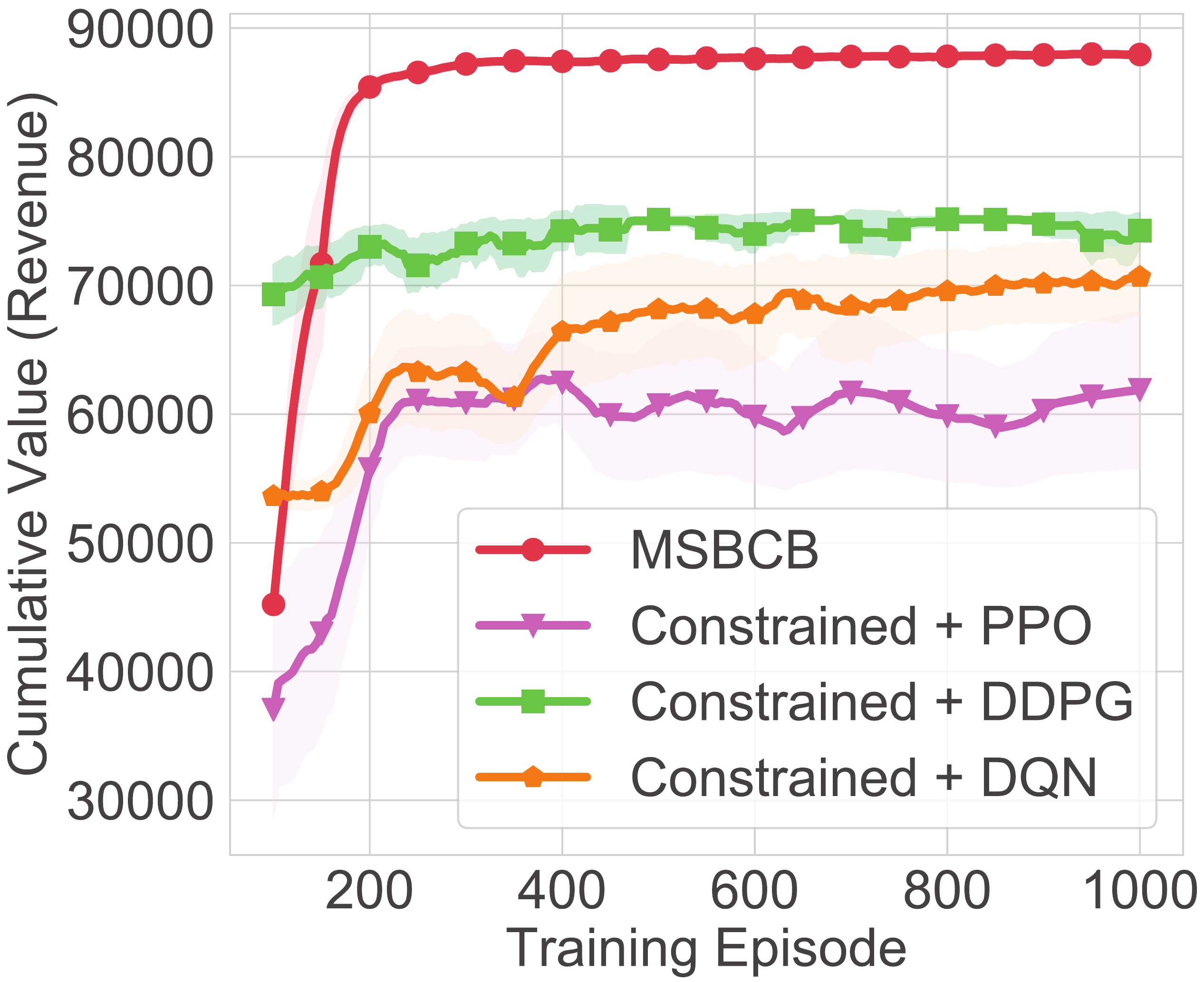}
  \caption{Values comparison (learning curves) of MSBCB and state-of-the-art CMDP based RL approaches.}
  \label{exp_sub_figure_3}
\end{figure}

\textbf{Decomposed MSBCB vs Undecomposed optimization.} Similar to \cite{wu2018budget}, the undecomposed optimization baselines consider all users’ requests as a whole and model the budget allocations among all request as a CMDP. As shown in Figure \ref{exp_sub_figure_3} and Table \ref{tab:all_method_results}, {MSBCB} outperforms the CMDP based RL approaches by a large margin. The reason of the poor performance in CMDP-based approaches is that these methods model all users' requests as a whole sequence and thus the learning process is particularly inefficient. In contrast, our {MSBCB} decomposes the whole sequence optimization into an efficient two-level optimization process, thus can achieve better performance more easily.
\begin{table}[htbp]
  \caption{Cumulative values, costs, value improvements (over {Contextual Bandit}) and the approximation ratio of all approaches.}
  \label{tab:all_method_results}
  \scalebox{0.63}{
   \begin{tabular}{lcccc}
     \toprule
     Method    & Revenue   & Cost & Revenue Impro & Approximation Ratio  \\
     \midrule
     Manual Bid        &  38838.28 & 11995.10 & -48.31\% & 43.5\%             \\
     Contextual Bandit     &  75137.30 & 11995.46 & 0\% & 84.15\%            \\
     \midrule
     Constrained + PPO  &  61890.92 & 11954.07 & -17.63$\pm$16.11\% & 69.31$\pm$13.56\%            \\
     Constrained + DDPG  &  74259.12 & 11996.12 & -1.19$\pm$3.66\% &  83.17$\pm$3.08\%           \\
     Constrained + DQN   &  70662.65 & 11881.12 & -5.96$\pm$7.83\% & 79.14$\pm$6.59\%             \\
     \midrule
     Greedy + maxCPR &  83668.70 & 11914.12 & 11.35$\pm$2.84\% & 93.70$\pm$2.36\%             \\
     \midrule
     Greedy + PPO        &  76970.35 & 11825.59 & 2.44$\pm$3.52\% & 86.20$\pm$2.93\%             \\
     Greedy + DDPG       &  80424.69 & 11841.28 & 7.04$\pm$1.13\% & 90.07$\pm$0.92\% \\
     Greedy + DQN        &  84117.09 & 11794.24 & 11.95$\pm$4.96\% & 94.21$\pm$4.14\%             \\
     \textbf{MSBCB}     &  \textbf{87947.99} & \textbf{11957.57} & \textbf{17.95}$\pm$\textbf{0.42}\% & \textbf{98.50}$\pm$\textbf{0.33}\%             \\
     \textbf{MSBCB (enum)}     &  \textbf{89251.77} & \textbf{11988.36} & \textbf{18.78}\% & \textbf{99.96}\%             \\
     \midrule
     Offline Optimal     &  89291.11 & 11999.23 & 18.84\% & 100.00\%              \\
     \bottomrule
   \end{tabular}
  }
\end{table}

The complete comparisons of all approaches are shown in Table \ref{tab:all_method_results}. The budget constraint $B$ is set to 12000 for all experiments. In Table \ref{tab:all_method_results},we also add an {MSBCB (enum)}, which is the theoretical upper bound of our {MSBCB}. The difference between {MSBCB (enum)} and {MSBCB} is that: the {MSBCB (enum)} computes the optimal advertising policy $\pi_i^*$ for each user $i$ by enumerating all possible policies. Instead of utilizing the RL approach, {MSBCB (enum)} could find the one which maximizes $V_{G}(i|\pi_i) - \text{CPR}_\text{thr}*V_{C}(i|\pi_i)$.
We see {MSBCB (enum)} is very close to the optimal solution and reaches an approximation ratio of 99.96\%.

\subsection{Effectiveness of Action Space Reduction}
\label{Effectiveness of Action Space Reduction}
As shown in Table \ref{tab:action space reduction}, {MSBCB} achieves a revenue of 75000 in only 61 epochs, reducing more than 60\% samples compared with the state-of-the-art RL baselines without using the action-space reduction technique. As for learning process, our {MSBCB} achieves the same revenue (80000) more than 10 times faster than the baselines, reducing more than 90\% samples and finally reaches the highest revenue. Thus, with the action space reduction technique, our {MSBCB} could reach a higher performance with a faster speed and significantly improve the sample efficiency.
More analysis of our {MSBCB}, e.g., the convergence of $\Pi^*$ and $\text{CPR}_\text{thr}^*$, and the hyperparameter settings of the offline experiments are shown in Section \ref{offline_env} of the Appendix.
\begin{table}[htbp]
  \caption{The training epochs and the number of samples needed by different approaches when achieving the same revenue level.}
  \label{tab:action space reduction}
  \scalebox{0.655}{
  \begin{tabular}{l|cc|cc|cc}
    \toprule
    Revenue & \multicolumn{2}{|c}{75000} & \multicolumn{2}{|c}{80000} & \multicolumn{2}{|c}{85000} \\
    Method     & \#Epoch & \#Samples & \#Epoch & \#Samples & \#Epoch & \#Samples \\
    \midrule
    Greedy+PPO          & 817 & 4183040 & - & - & - & -        \\
    Greedy+DDPG         & 154 & 788480  &  853 & 4362240  & - & -      \\
    Greedy+DQN          & 373 & 1909760  & 754 & 3855360  & - &-          \\
    \textbf{MSBCB}               & \textbf{61} & \textbf{312320}   & \textbf{71} & \textbf{363520}    & \textbf{104} & \textbf{532480}        \\
    \bottomrule
  \end{tabular}}
\end{table}

\section{Empirical Evaluation: Online A/B Testing}
\begin{figure}[htbp]
   \centering
   \includegraphics[width=3.25in,height=1.3in]{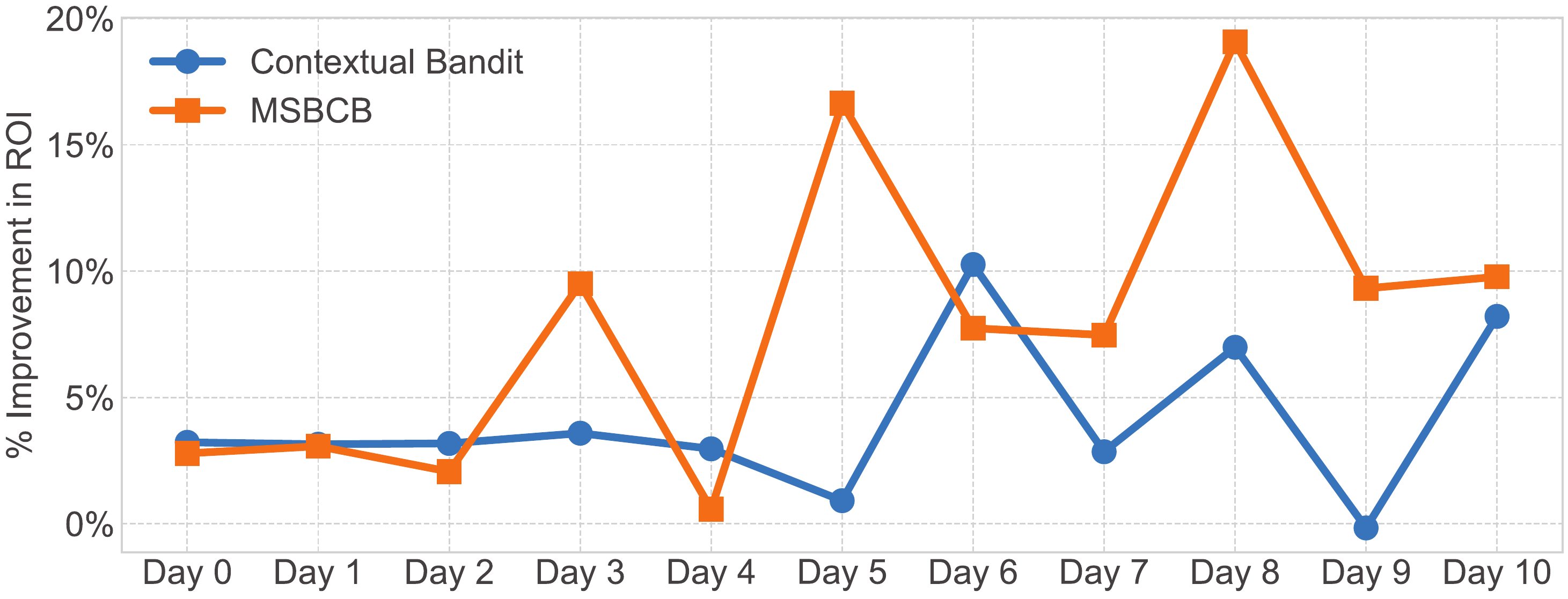}
   \caption{Daily ROI improvement comparisons of {Contextual Bandit} and {MSBCB} over {Cross Entropy Method}.}
   \label{fig:daily_results}
\end{figure}

We deployed MSBCB on one of the world's largest E-commerce platforms, Taobao. Our platform is authorized by the advertisers to dynamically adjust their bid prices for each user request according its value in the real-time auction. In the online experiments,
we compare {MSBCB} with two models widely used in the industry.
\begin{itemize}
  \item {Cross Entropy Method} ({CEM}), which is a deployed production model, whose target is to optimize the immediate rewards. We consider {CEM} as the control group in the following evaluations.
  \item {Contextual Bandit}, which has been explained in previous section and is reserved as a contrast test.
\end{itemize}
The experiment involves 135,858,118 users and 72,147 ad items from 186 advertisers. For fair comparison, we control the consumers and the advertisers involved in the A/B testing to be homogeneous. In detail, the 135,858,118 users are randomly and evenly divided into 3 groups. For users in group \#1, all 186 advertisers adopt the {CEM} algorithm. For users in group \#2, all 186 advertisers adopt the {Contextual Bandit} algorithm. For users in group \#3, all 186 advertisers adopt our {MSBCB}. Table \ref{fig:live_results} summarises the effects of the {Contextual Bandit} and our {MSBCB} compared to the {Cross Entropy Method} from Dec.10 to Dec.20 in 2019. From Table \ref{fig:live_results}, we see that our {MSBCB} achieves a +10.08\% improvement in revenue and a +10.31\% improvement in ROI with almost the same cost (-0.20\%). The results indicate that upgrading the myopic advertising strategy into a farsighted one could significantly improves the cumulative revenue. Besides, as shown in Figure \ref{fig:daily_results}, the daily ROI improvement also demonstrates the effectiveness of our {MSBCB} compared with the {Contextual Bandit}.

\begin{table}[htbp]
   \caption{
    The overall performance comparisons of the A/B testing.
    CVR represents the Conversion Rate of the users. \#PV represents the number of page views. $\text{ROI}=\frac{\text{Revenue}}{\text{Cost}}$ means Return On Investment. (Notice that {CEM} is the control group and the improvements of {Contextual Bandit} and {MSBCB} are compared over {CEM}.)
   }
   \label{fig:live_results}
   \scalebox{0.735}{
   \begin{tabular}{c | c c  c  c  c }
      \toprule
      Method    & Revenue & Cost & CVR & \#PV & ROI \\
      \midrule
      Contextual Bandit     & +0.91\%  & -3.26\%  &   +4.78\% &  +4.62\%  &  +4.31\% \\
      \textbf{MSBCB}    & \textbf{+10.08}\%  & \textbf{-0.20}\%  & \textbf{+6.04}\% & \textbf{+15.37}\%  &  \textbf{+10.31}\%\\
      \bottomrule
   \end{tabular}
   }
\end{table}

Given that there are only 186 advertisers take part in our online experiment, one frequently asked question is``\textbf{How does the MSBCB work across all ads?}" Since 186 is relatively small compared with the total number of advertisers, their policy updates would not cause dramatic changes to the RTB environment. In other words, the RTB environment is still approximately stationary from a single-ad perspective. This setting also works well with our practical business model–-providing better service for VIP advertisers (about 0.2\% of all the advertisers). In the case that the majority of the advertisers adopt MSBCB, the system cannot be estimated as being stationary from any single-ad’s perspective and explicit multi-agent modeling and coordination should be incorporated. Detailed analysis of the improvement in revenue for each advertiser is presented in Table \ref{tab:shop_results} and Figure \ref{fig:roi_details} of the Appendix. More details about the deployment and experimental results (e.g., the online model architecture) can also be found in Section \ref{deployment} and \ref{online_ab_testing} of the Appendix.

\section{Conclusion}
We formulate the multi-channel sequential advertising problem as a {Dynamic Knapsack Problem}, whose target is to maximize the long-term cumulative revenue over a period of time under a budget constraint. We decompose the original problem into an easier bilevel optimization, which significantly reduces the solution space. For the lower-level optimization, we derive an optimal reward function with theoretical guarantees and design an action space reduction technique to improve the sample efficiency.
Extensive offline experimental analysis and online A/B testing demonstrate the superior performance of our {MSBCB} over the state-of-the-art baselines in terms of cumulative revenue.

\section*{Acknowledgements}
The work is supported by the National Natural Science Foundation of China (Grant Nos.: 61702362, U1836214), the Special Program of Artificial Intelligence and the Special Program of Artificial Intelligence of Tianjin Municipal Science and Technology Commission (No.: 569 17ZXRGGX00150) and the Alibaba Group through Alibaba Innovative Research Program. We deeply appreciate all teammates from Alibaba group for the significant supports for the online experiments.

\bibliography{seq_advertising}
\bibliographystyle{icml2020}

\appendix

\setcounter{table}{3}
\setcounter{figure}{8}
\setcounter{equation}{12}
\renewcommand{\algorithmicrequire}{ \textbf{Policy Evaluation:}} 
\renewcommand{\algorithmicensure}{ \textbf{Policy Improvement:}} 
\newcommand{\algorithmicinit}{\textbf{Initialize:}}
\newcommand{\INIT}{\item[\algorithmicinit]}
\newcommand{\algorithmicreturn}{\textbf{Return:}}
\newcommand{\RETURN}{\item[\algorithmicreturn]}

\icmltitlerunning{Appendix}

\onecolumn
\icmltitle{Appendix}

\begin{appendices}
\section{Background of Online Advertising}
Online advertising is a marketing strategy involving the use of \textit{advertising platform} as a medium to obtain website traffics and targets, and deliver marketing messages of \textit{advertisers }to the suitable \textit{customers}.

\textit{Platform.} Advertising platform plays an important role in connecting consumers and advertisers. For consumers, it provides multiple advertising channels, e.g., channels on news media, social media, E-commerce websites and apps to explore. For advertisers, it provides automated bidding strategies to compete for consumers in all channels under the setting of real-time bidding (RTB), in which advertisers bid for ad exposures and the exposures opportunities go to the highest bidder with a cost which equals to the second-highest bid in the auction.

\textit{Consumers.} Consumers explore multiple channels during the several visits to the platform within a couple of days. A consumer's final purchase of an item is usually a gradually changing process, which often includes the phases of Awareness, Interest, Desire, and Action (AIDA) \cite{roberge2015sales}. The consumer's decision to convert (purchase a product) is usually and has to be driven by multiple touchpoints (exposures) with ads. Each advertising exposure during the sequentially multiple interactions could influence the consumer’s mind (preferences and interests) and therefore contribute to the final conversion.

\textit{Advertisers.} The goal of advertisers is to cultivate the consumer's awareness, interest and finally driving purchase. As different ad strategies can affect consumers' AIDA, an advertiser should develop a competitive strategy to win the ad exposures in RTB setting. When the ad is displayed to a consumer, in Cost Per Click (CPC) setting, the advertisers should pay commission to the platform after the consumer clicking the ad. When the consumer purchases the advertised item, the advertiser will get the corresponding revenue.

The objective of an advertiser is usually to optimize the accumulated revenue within a time period under a budget constraint. A strategy that maximizes short-term revenue of each ad exposure on different channels independently is obviously unreasonable, since the final purchase is a result of long-term ad-consumer sequential interactions and the consumer's visits between different channels are interdependent. Therefore, the advertiser must develop a strategy to overcome following two key challenges: (1) Find the optimal interaction sequence including interaction times, channels selection and channel orders for a targeted consumer; (2) Choose targeted consumers and allocate predefined limited budget to them in multiple interaction sequences.

\begin{figure*}[htbp]
\centering
\includegraphics[height=1.6in, width=6.5in]{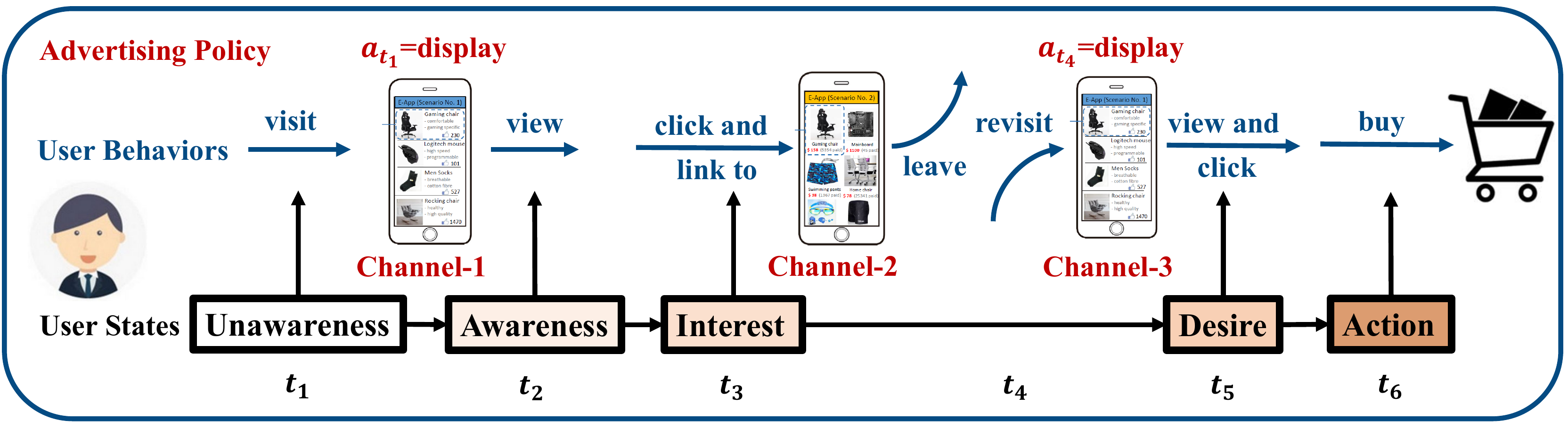}
\caption{An illustration of the sequential multiple interactions (across different channels) between a user and an ad. Each ad exposure has long-term influence on the user's final purchase decision.}
\label{Figure:user-journey}
\end{figure*}

An example of a user's shopping journey is shown in Figure \ref{Figure:user-journey}. At time $t_1$, a user visits the news media channel and triggers an advertising exposure opportunity. Then, the advertising agent executes a display action and leaves an exposure on the user. After that, the user becomes aware of and is interested in the commodity, so he clicks the hyperlink. Quickly, the user is induced into the landing (detail) page of the commodity in the shopping app. After fully understanding the product information, the user leaves the shopping app. After a period of time, the user comes back to the shopping app at time $t_4$ and triggers an exposure opportunity of banner advertising. The advertising agent executes a display action as well. Consequently, the user’s desire is stimulated. At time $t_6$, the user makes a purchase. In this example, the ad exposure at time $t_1$ influences the user’s mind and contributes to the ad exposure at time $t_4$ and the delayed purchase, which means
the ad exposure on one channel would influence the user’s preferences and
interests, and therefore contributes to the final conversion. Thus, the goal of advertising should maximize the total cumulative revenue over a period of time instead of simply maximizing the immediate revenue.

\section{Proof and Analysis}
\label{proof}
\subsection{Knapsack Problem in Online Advertising Settings}
\label{proof_greedy}
\textbf{Theorem 2.} \emph{The greedy solution to the proposed dynamic knapsack problem of online advertising is $\lambda $ approximately optimal where $\lambda > 99.9\%$ }

\emph{Proof.} In the proposed online advertising problem, each user is with value $V_{G}$ (i.e. the profit of advertiser when the user purchase the commodity) and weight $V_{C}$ (i.e. the total budget consumption for the target user in the real-time bidding to reach the final purchase). As the item (i.e. user) is non-splittable, the proposed dynamic knapsack problem is essentially a 0-1 knapsack problem which aims to maximize the total value of the knapsack given a fixed capacity $B$. For each item, we can calculate the Cost-Performance Ratio (CPR) as $V_{G}/V_{C}$. Sort all items in descending order of CPR, i.e. $\left(V_{G_1}, V_{C_1}\right),\left(V_{G_2}, V_{C_2}\right), \dots,\left(V_{G_n}, V_{C_n}\right)$ where $\text{CPR}_{i} \geq \text{CPR}_{j}, \forall i \leq j \leq n$. For $V_{C} > 0$, $V_{G} > 0$ and $B > 0$, we first define that this 0-1 knapsack problem has optimal solution $K^{*}(V_{C}, V_{G}, B)$ and greedy solution $K(V_{C}, V_{G}, B)$ where $K^{*}$ and $K$ represent the total value of the knapsack.

Assume $B_{end}$ is the remaining budget after greedy algorithm, the following inequality holds:
\begin{equation}
\label{greedy solution derivation}
    \frac{B-B_{e n d}}{B} K^{*}(V_{C}, V_{G}, B) \leq K(V_{C}, V_{G}, B) \leq K^{*}(V_{C}, V_{G}, B)
\end{equation}

This is because:
\begin{enumerate}[1)]
\item If the knapsack can hold all the items after the greedy algorithm, that is, the optimal solution is equal to the greedy solution. As $B_{end} \ge 0$, we have $
\frac{B-B_{e n d}}{B} K^{*}(V_{C}, V_{G}, B) \leq K^{*}(V_{C}, V_{G}, B)=K(V_{C}, V_{G}, B)$
\item If the knapsack cannot hold all the items after the greedy algorithm, as $\frac{V_{G_1}}{V_{C_1}} \ge \frac{V_{G_2}}{V_{C_2}} \ge ... \ge \frac{V_{G_l}}{V_{C_l}} $, we have $V_{G_l} \sum_{j=1}^{l-1} V_{C_j} \leq V_{C_l} \sum_{j=1}^{l-1} V_{G_j} \Leftrightarrow V_{G_l}\left(B-B_{e n d}\right) \leq V_{C_l} K(V_{C}, V_{G}, B) \Leftrightarrow \frac{V_{G_l}}{V_{C_l}} \leq \frac{K(V_{C}, V_{G}, B)}{B-B_{e n d}} \Leftrightarrow K(V_{C}, V_{G}, B) \geq K^{*}(V_{C}, V_{G}, B)-\frac{B_{e n d} K(V_{C}, V_{G}, B)}{B-B_{e n d}}$ where $l$ is the index of last item picked by greedy algorithm. This derivation can be simplified to $K(V_{C}, V_{G}, B) \geq \frac{B-B_{e n d}}{B} K^{*}(V_{C}, V_{G}, B)$.
\end{enumerate}

In online advertising settings, the budget spent on a single user is much smaller than the advertiser's total budget. We conduct statistics on one of the world's largest E-commerce platforms to prove it. On Feb 3rd of 2020, a total of 1136149 ads result in 983414548 user-ad sequences (a user sequence consists of multiple interactions of the same user with the same ad), with an average of 865 user sequences per ad.
Interactions with users of each ad forms a knapsack problem, where each user sequence is an item in the knapsack.
The average maximum budget consumed by each user sequence accounts for 0.07068\% of the total budget capacity of the advertisers.
We also list details of 5 ads with largest budget consumption in Table \ref{detail cost and budget}, where the maximum budget consumed by each user sequence is much smaller than 1/1000 (smaller than 3/10000 specifically) of the total budget of each ad.

\begin{table*}[htbp]
   \centering
   \resizebox{0.9\columnwidth}{!}{
   \begin{tabular}{c | c c  c  c  c c}
      \toprule
      Ad      & \#Users Sequences & Budget & Avg Cost & (Avg Cost)/Budget & Max Cost & (Max Cost)/Budget \\
      \midrule
      Ad 1       & 2460976 &	119352.51 &	0.048498039 & 0.0000406343\% & 20.04 & 0.0167905979\% \\
      Ad 2       & 2674738 &	114388.54 &	0.04276626 & 0.000037388\% &	26.22 &	0.0229218766\% \\
      Ad 3       & 2848816 &	90113.08 &	0.031631766 &	 0.0000351023\% &	15.29&	0.0169675701\% \\
      Ad 4       & 2107497 &	82951.82 &	0.03936035	& 0.0000474497\% &	5.6	     &  0.0067509067\% \\
      Ad 5       & 1087011 &	77140.49 &	0.070965694	& 0.0000919954\% &	19.32	 &   0.0250452130\% \\
      \bottomrule
   \end{tabular}%
   }
   \caption{Detailed Comparison between an ad's total budget and cost on a user sequence.}
   \label{detail cost and budget}
\end{table*}

As proposed in \citet{dantzig1957discrete}, $\forall i \in 1,2, \ldots, n, V_{C_i} \leq(1-\lambda) B, 0 \leq \lambda \leq 1$, the greedy algorithm achieves an approximation guarantee of $\lambda$. We can conclude from above statistics that $\max_i \frac{V_{C_i}}{B}\!\le\!\frac{1}{1000}$, which means $\lambda$ is much greater than $1\!-\!\frac{1}{1000}$.

The thesis above can be further proved:
\begin{enumerate}[1)]
\item If the knapsack can hold all the items after the greedy algorithm, that is, the greedy solution is obviously equal to the optimal solution, which is also the $\lambda$ approximately optimal solution.
\item If the knapsack cannot hold all the items after the greedy algorithm, we have $V_{C_l}>B_{e n d}$, that is, $B_{e n d}<V_{C_l} \leq(1-\lambda) B$. According to Formula \ref{greedy solution derivation}, we have
\begin{equation}
    \begin{aligned} K(V_{C}, V_{G}, B) & \geq \frac{B-B_{e n d}}{B} K^{*}(V_{C}, V_{G}, B) \\ &>\frac{B-(1-\lambda) B}{B} K^{*}(V_{C}, V_{G}, B) \\ &=\lambda K^{*}(V_{C}, V_{G}, B) \end{aligned}
\end{equation}
\end{enumerate}

Therefore, in theory, the greedy solution in our online advertising settings is $\lambda$ approximately optimal and the $\lambda$ is much greater than 99.9\% in our case.

\subsection{Regretless Optimal Bidding Strategy $b^*_t$}
\label{proof_optimal_bid}
\textbf{Theorem 3.} \emph{During the online bidding phase, the bidding agent can always set the bid price as:}
\begin{equation}
\begin{aligned}\textbf{b}_t^*=&\left[\left(\frac{Q_G(s,\widehat{a_t}=1)}{\text{CPR}_\text{thr}^{*}}-Q_C^{\text{next}}(s,\widehat{a_t}=1)\right)\right.\\
-&\left.\left(\frac{Q_G(s,\widehat{a_t}=0)}{\text{CPR}_\text{thr}^{*}}-Q_C^{\text{next}}(s,\widehat{a_t}=0)\right)\right]
\end{aligned}
\end{equation}
\emph{where $Q_C^{\text{next}}(s,\widehat{a_t})=\mathbb{E}[\sum_{k=t+1}^{T_j}c_k|s,\widehat{a_t},\pi_j]$. $\textbf{b}_t^*$ is a regretless optimal bidding strategy without any loss of accuracy.  }

\emph{Proof.} Since $\textbf{bid}_{t}^{\textbf{2nd}}$ is unknown until the current auction is finished, we prove the regretless of $\textbf{b}_t^*$ from the following two cases:
\begin{enumerate}[1)]
\item If $\textbf{b}_t^*\!>\!\textbf{bid}_{t}^{\textbf{2nd}}$: $\textbf{b}_t^*\!>\!\textbf{bid}_{t}^{\textbf{2nd}}\!\Leftrightarrow\! Q(s,\widehat{a_t}\!=\!1)\!>\!Q(s,\widehat{a_t}\!=\!0)$, which means the agent should take action $\widehat{a_t}=1$ in this case. Exactly, $\textbf{b}_t^*$ is greater than the second highest price $\textbf{bid}_{t}^{\textbf{2nd}}$ based on the condition for entering the current branch. Thus, the agent will always win the auction and the executed action is indeed $\widehat{a_t}=1$.
\item If $\textbf{b}_t^* \le \textbf{bid}_{t}^{\textbf{2nd}}$: $\textbf{b}_t^*\!\le\!\textbf{bid}_{t}^{\textbf{2nd}}\!\Leftrightarrow\! Q(s,\widehat{a_t}\!=\!1)\!\le\!Q(s,\widehat{a_t}\!=\!0)$, which means the agent should take action $\widehat{a_t}=0$ in this case. Exactly, $\textbf{b}_t^*$ is less than the second highest price $\textbf{bid}_{t}^{\textbf{2nd}}$ according to the condition. Thus, the agent will always lose the auction and the executed action is indeed $\widehat{a_t}=0$.
\end{enumerate}
Thus, we complete the proof.

\subsection{Convergence Analysis of \emph{MSBCB}}
\label{proof_convergence}
\begin{figure}[htbp]
  \centering
  \includegraphics[scale=0.6]{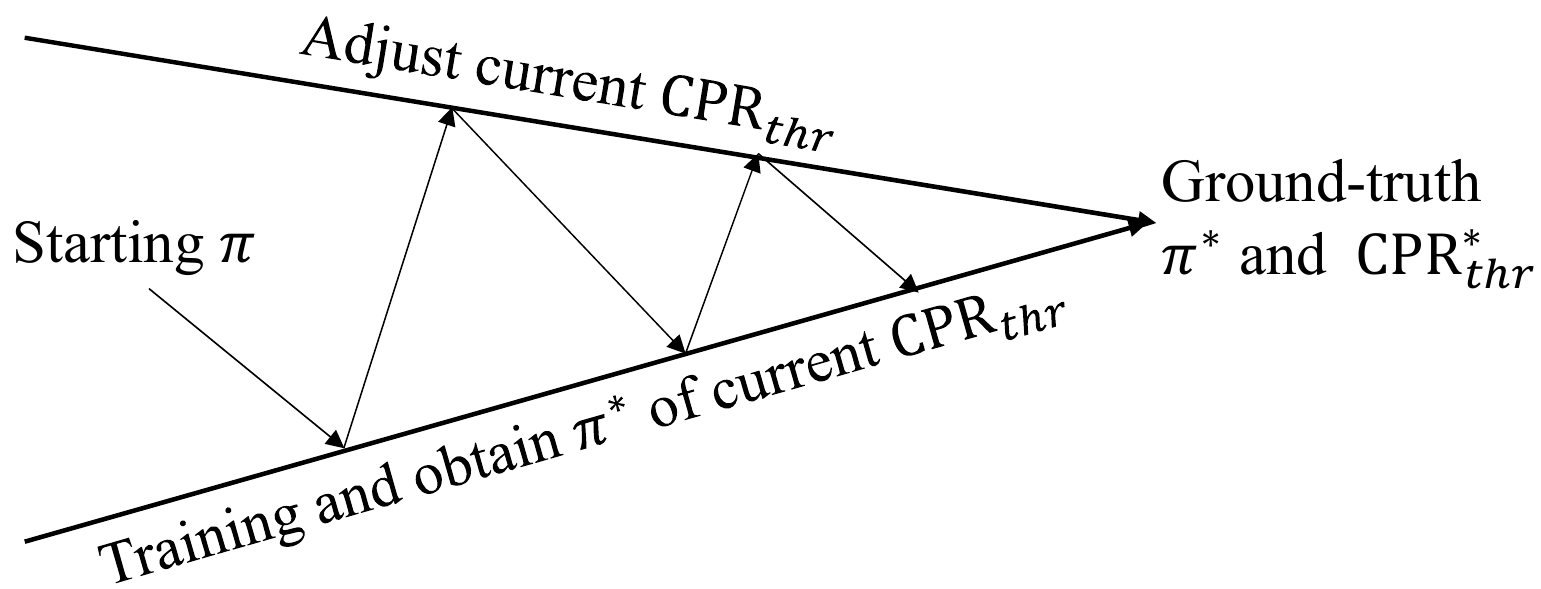}
  \caption{Convergence demonstration of \emph{MSBCB}}
  \label{convergence demonstration}
\end{figure}
The overall framework of \emph{MSBCB} can be described as follows:
\begin{enumerate}[(1)]
  \item Let the budget constraint of an advertiser be $B$. Given a $\text{CPR}_\text{thr}$, we can use reinforcement learning algorithms to ensure that each user $i$ is optimized according to $\pi_i^* \coloneqq \mathop{\text{argmax}}_{\pi_i} \left[V_{G}(i|\pi_i) - \text{CPR}_\text{thr}*V_{C}(i|\pi_i)\right]$ and converges to the optimal policy $\pi_i^*$ under the current $\text{CPR}_\text{thr}$. Further, picking all users whose $\text{CPR}_i \ge \text{CPR}_\text{thr}$ will result in a total cost of $B'$ (i.e., the advertiser spends a budget $B'$).
  \item As the current estimated threshold $\text{CPR}_\text{thr}$ might have some bias from the optimal $\text{CPR}_\text{thr}^*$, $B'$ may not equal to the budget $B$. Thus, we design a PID controller to dynamically adjust the estimated $\text{CPR}_\text{thr}^*$ so as to minimize the gap between the budget constraint $B$ and the actual feedback of the daily cost $B'$.
\end{enumerate}
As described in Figure \ref{convergence demonstration}, \emph{MSBCB} repeats the above two steps iteratively. Given an updated $\text{CPR}_\text{thr}$, each $\pi$ will be optimized by the lower-level reinforcement learning algorithms and $\pi$ will move towards the optimal $\pi^*$. As a result, users whose optimized $\text{CPR}_i \ge \text{CPR}_\text{thr}$ will be selected and we get the daily cost $B'$. Then, the current $\text{CPR}_\text{thr}$ will be updated so that the gap between the cost $B'$ and the budget $B$ will be further minimized.
Thus, $\text{CPR}_\text{thr}$ will move towards the optimal $\text{CPR}_\text{thr}^*$ gradually.
As long as the learning rates of $\pi$ and $\text{CPR}_\text{thr}$ are small enough, the overall iterations will finally converge. In this paper, we also validate the convergence of our \emph{MSBCB} in the experiments. As shown in Section \ref{offline_results} of the paper, our method converges quickly and finally reaches an approximation ratio of 98.53\%.

\section{Deployment}
\label{deployment}
Here we give the online deployment details of our \emph{MSBCB}.
\subsection{Myopic to Non-Myopic Advertising System Upgrade Solution}

\begin{figure*}[htbp]
  \centering
  \includegraphics[scale=0.7]{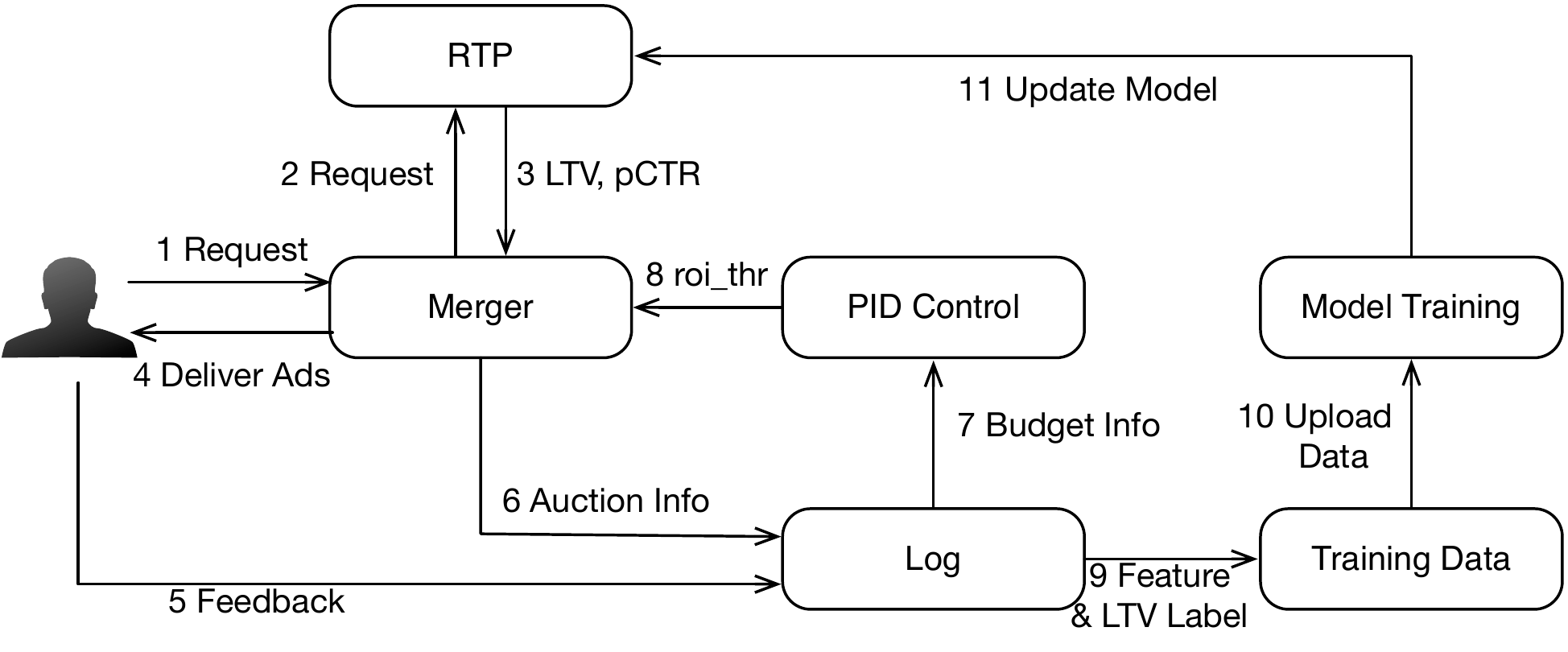}
  \caption{Online System}
  \label{online-system}
\end{figure*}

A myopic advertising system includes several key components as Figure \ref{online-system} shows: (1) Log module collects auction information and user feedback. (2) Training data are constructed based on log followed by model training with offline evaluation. (3) Real-time prediction (RTP) module provides service for myopic value prediction of user-ad pairs. RTP periodically pulls newly trained models. (4) Merger module receives the user visit, requests RTP for myopic value with which ad bid adjustment ratios and ranking scores are calculated (In advertising, ranking score is $ecpm=pCTR * bid$ where $pCTR$ is predicted Click Through Rate and $bid$ is the bidding price). Finally, top-scored ads are delivered to the user. Above myopic advertising system can upgrade to a non-myopic system by considering the following key changes. (1) Log module needs to keep long-term auction information and users' feedback, and these data are used to construct features and long-term labels for training. Besides, logged data have to track each advertised item's budget and current cost data which are fed to a PID control module to compute $\text{CPR}_\text{thr}$ for users selection in Merger. (2) Model training can use Monte Carlo (MC) or Temporal Difference (TD) methods. For MC, the long-term labels are cumulative rewards of a sequence and the training becomes a supervised regression problem. For TD, one-step or multi-step rewards are used to compute a bootstrapped long-term value using a separate network for training. (3) RTP module should periodically pull both myopic and non-myopic newly trained models and provide corresponding value prediction service. (4) Merger maintains an $<\text{item}, \text{CPR}_\text{thr} >$ table which is updated periodically from PID module. When a user visit comes, Merger requests RTP for both $pCTR$ and long-term values (long-term $GMV$ i.e. $V_{G}$ and $cost$ i.e. $V_{C}$ in our paper), and with $\text{CPR}_\text{thr}$ decides the selection of current user and bid adjustment.

\subsection{Long-Term Value Prediction Model}
\begin{figure*}[htbp]
  \centering
  \includegraphics[width=6.6in]{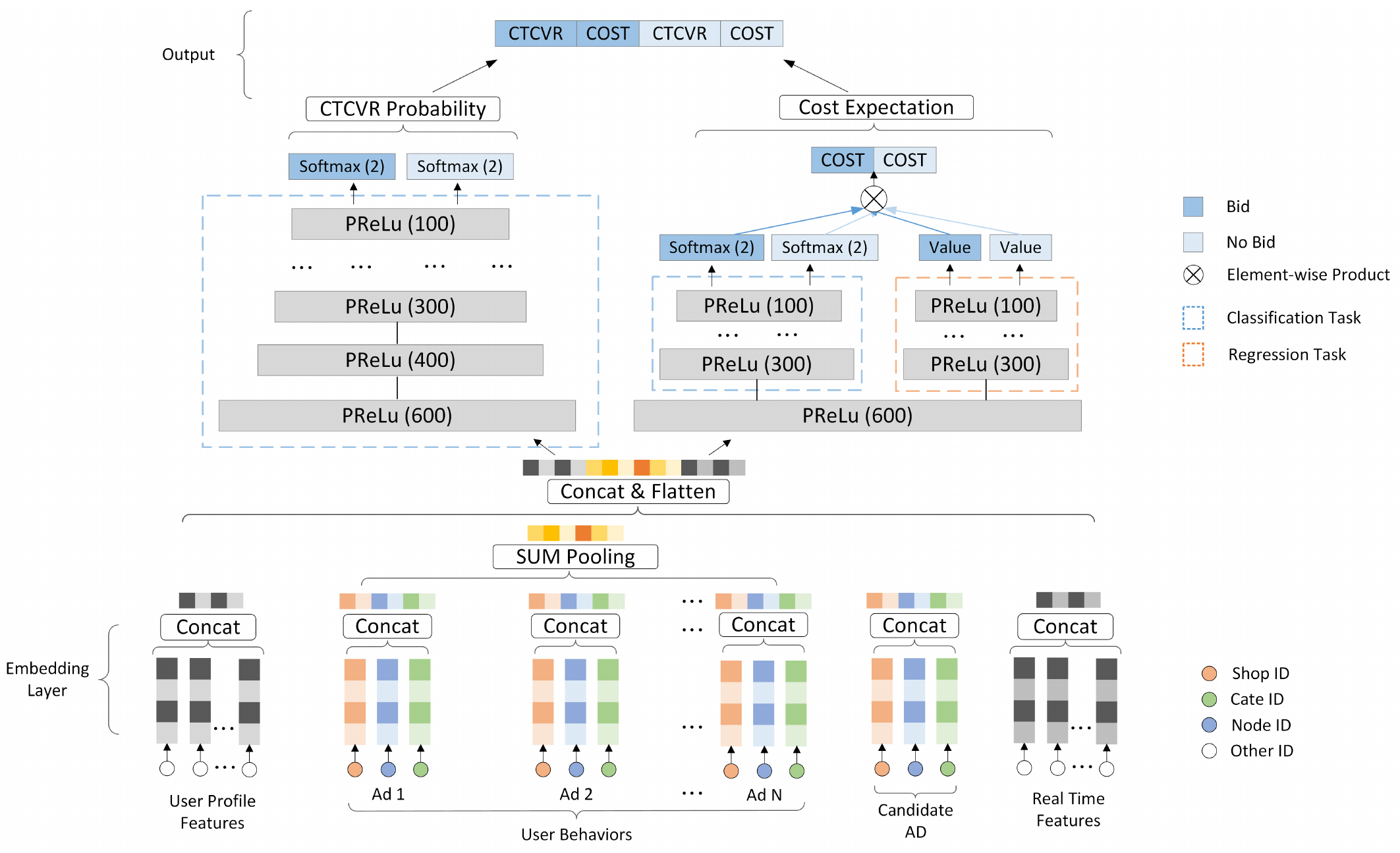}
  \caption{Long-Term Value Prediction Model}
  \label{LTV-prediction-model}
\end{figure*}

\subsubsection{Features and Labels}
Features for long-term value prediction should contain sufficient user's static profile and historical behavior information. Most myopic advertising systems already have a sound feature system which can summarize user-oriented, ad-oriented and user-ad interactive history very well. Besides, due to the large amount of data collected by the online advertising system, these features are able to generalize across large number users where each user-ad pair's interaction is considered as a separate MDP, thus, help the prediction model learning.
To be specific, the state $s_t$ at step $t$ includes: 1) user profile features; 2) user behavior features; 3) real-time user behavior features; 4) context features; 5) user-ad interaction histories; 6) user feedback before current step $t$ and so on. Features are constructed based on past 7-14 days data before user visit time t. For the MC training method, labels are constructed using the following 7 days data after user visit time $t$. For the TD method, labels are the instant rewards at time $t$ and the long-term labels are constructed using a bootstrap method.

\subsubsection{Model Architecture}
The long-term value model architecture is shown in Figure \ref{LTV-prediction-model}, where the model takes the features as input and output long-term value of $GMV$ (i.e. $V_{G}$ in our formulation) and $cost$ (i.e. $V_{C}$ in our formulation) for both action=1 (display the ad) and action = 0 (do not display the ad).

We use one model to output multiple long-term values ($GMV$ and $cost$ for action=1 and action=0). Multiple prediction tasks share the same bottom layers because we consider the underlying knowledge of the user's sequence behaviors such as opening the app, jumping across channels, turning off the phone and revisiting the app should be learned together and shared. The shared layer converts input features to embeddings and embeddings in the same group are concatenated. The user-behavior group embeddings are then pooled with sum operation. User-profile embeddings, user-behavior embeddings, candidate ad embeddings, and real-time features are finally concatenated and flattened as the output of the bottom layers.

Following the shared bottom layers, the network is split into two forward-pass branches where one is for long-term $GMV$ prediction and one for long-term cost prediction. We find this two-branch design can reduce the influences among different tasks and stabilize the learning. For the long-term $GMV$ prediction, since each user usually buys a commodity only once, we only have to predict $P(buy>0|feature)$ denoted as $CTCVR$. In the online inference phase, the long-term $GMV$ is computed with $GMV=P(buy>0|feature) * item\_price$ where $item\_price$ is the price of the commodity. For the long-term cost prediction, in CPC (Cost-Per-Click) advertising, a user usually clicks several times before buys and the cost per click along with each click varies, thus, the long-term $cost$ prediction cannot be decomposed as the long-term $GMV$ prediction and the only way is to regress the long-term cost value. However, as most sequences' costs are zero, the direct regression learning process will be very noisy. Therefore, we design an additional hidden layer to compute $P(cost>0|feature),P(cost=0|feature)$ and $E(cost|cost>0, feature)$. Then, the predicted long-term cost is computed as $pcost = P(cost>0|feature)*E(cost|cost>0, feature)+P(cost=0|feature)*0=P(cost>0|feature)*E(cost|cost>0, feature)$ where $P(cost>0|feature)$ and $P(cost=0|feature)$ are learned using logistic regression loss and $pcost$ is learned using mean-square error loss $(pcost-cost)^2$. We find the above designs help improve the model's prediction performance in practice. For $CTCVR$ and $P(cost>0|feature)$, $P(cost=0|feature)$, we use GAUC \cite{zhou2018deep} as metric, and for $pcost$ regression, we use mean-square error and reverse order metrics.


\section{Empirical Evaluation: Supplementary of Offline Experiments}\
\label{offline_env}
\subsection{Experiments Settings.} Considering the potential losses of assets and money, it's usually forbidden to do a lot of trial and error and thoroughly comparisons between available baselines in a live advertising system. Thus we implement a fairly general simulation environment so that we could make extensive analyses of our approach. All experiments are conducted on an Intel(R) Xeon(R) E5-2682 v4 processor based Red Had Enterprise Linux Server, which consists of two processors (each with 16 cores), running at 2.50GHz (16 cores in total) with 32KB of L1, 256 KB of L2, 40MB of unified L3 cache, and 128 GB of memory and 2 Tesla M40 GPUs.
\subsection{Simulation Environment.} Here, we give the detail of the simulation environment. Similar to \cite{ie2019reinforcement}, the simulation environment includes the following 5 modules:
\vskip -0.3cm
\begin{itemize}
\setlength{\itemsep}{1.6pt}
\setlength{\parsep}{1.6pt}
\setlength{\parskip}{1.6pt}
  \item \emph{Advertisements and Topic Model:} We assume a set of documents $\mathcal{D}$ representing the content available for advertising. We also assume a set of topics (or users interests) $\mathcal{T}$ that capture fundamental characteristics of interest to users; we assume topics are indexed $1,2,...|\mathcal{T}|$. Each commodity $d \in\mathcal{D}$ has an associated topic vector $\mathbf{d}\in[0,1]^{|\mathcal{T}|}$, where $d_j$ is the degree to which $d$ reflects topic $j$. Each document $d \in\mathcal{D}$ also have an inherent quality ${Q}_d\in[0,1]$, representing the topic-independent attractiveness to the average user.
  \item \emph{Consumer Interest and Satisfaction Model:} Each user $i$ has various degrees of interests in topics, ranging from 0 (completely uninterested) to 1 (fully interested), with each user $i$ associated with an interest vector $\mathbf{u}\in[0,1]^{|\mathcal{T}|}$. Consumer $i$'s interest in advertisement $d$ is given by the dot product $I(u,d) = \mathbf{u}\mathbf{d}$. We assume some prior distribution $P_u$ over user interest vectors, but user $i$'s interest vector is dynamic, i.e., influenced by their advertisement consumption (see below). Besides, a user's satisfaction $S(u,d)$ with a consumed (viewed) advertisement $d$ is a function $f(I(u,d),{Q}_d)$ of user $i$'s interest and ad $d$'s quality. Here, we assume a simple convex combination $S(u,d)=(1-\alpha)I(u,d)+\alpha{Q}_d$. Satisfaction influences user dynamics as we discuss below.
  \item \emph{Consumer Choice Model:} The user's Click-Through Rate (CTR) and Conversion Rate (cvr) are represented by $I(u,d)$ and $S(u,d)$ respectively. Each user has the probability of clicking and buying an advertising commodity according the CTR and CVR.
  \item \emph{Consumer Dynamics:} We assume that a user's interest evolves as a function of the documents consumed (viewed). When user $i$ consumes document $d$, her interest in topic $T(d)$ is nudged stochastically, biased slightly towards increasing her interest, but allows some chance of decreasing her interest. In this paper, we set $\mathbf{u}\leftarrow\gamma\mathbf{u}+\beta*S(u,d)*\mathbf{d}$, where $\gamma$ is the interest decay rate and $\beta\in[-1,1]$ is a user independent parameter.
  \item \emph{Consumer Visiting Model and Advertising System Dynamics:} The users' request sequence are generated from a stable distribution $P_{req}$. For each user's request, all  advertisements $d\in\mathcal{D}$ give a bid and competes with other bidders in real-time. The winner has the privilege to display its ad to the user, which could further influence the user's interest and behavior.
\end{itemize}
\vskip -0.35cm
\subsection{Codes and Datasets.} The codes and datasets to reproduce our offline experiments are provided in another supplementary material.

\subsection{Cost Comparison.}
\begin{figure}[ht]
  \centering
  \includegraphics[width=3.2in]{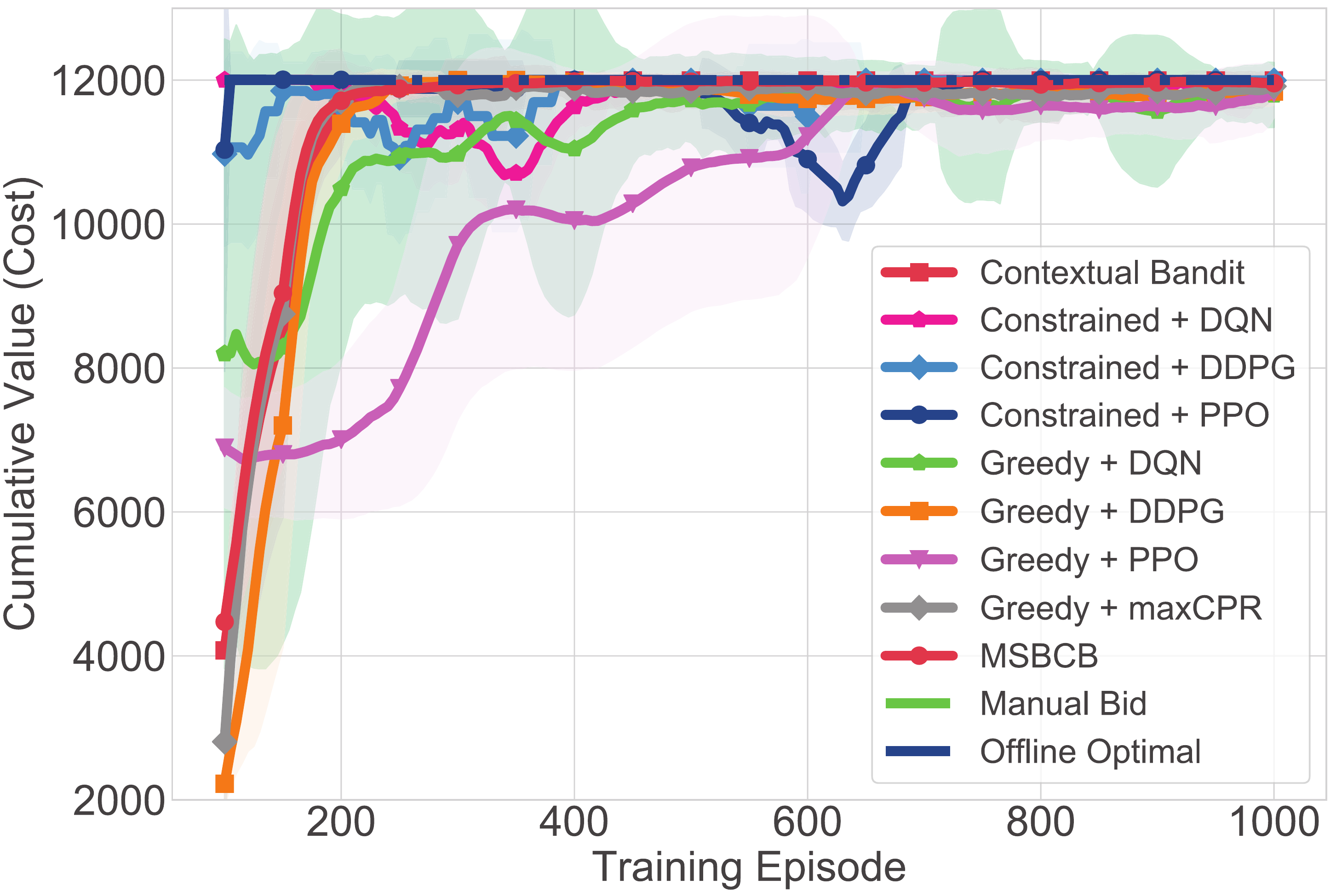}
  \caption{The learning curves of costs of our \emph{MSBCB} and the other baseline approaches.}
  \label{exp_cost_figure}
\end{figure}

The consumption of budget during the training process is shown in Figure \ref{exp_cost_figure}. As we can see, the costs of all approaches converge to about 12000, which is exactly equal to the budget we set in experiments. Specific costs of each approach can be found in Table 1 of paper.

\subsection{Convergence Analyses}
\subsubsection{Convergence of each $\pi_i^*$ given any $\text{CPR}_\text{thr}$.} 
\begin{figure*}[htb]
\centering
\subfigure[user \#1's learned policy.] {
\label{Figure:case-1-policy}
\includegraphics[width=2.1in,angle=0]{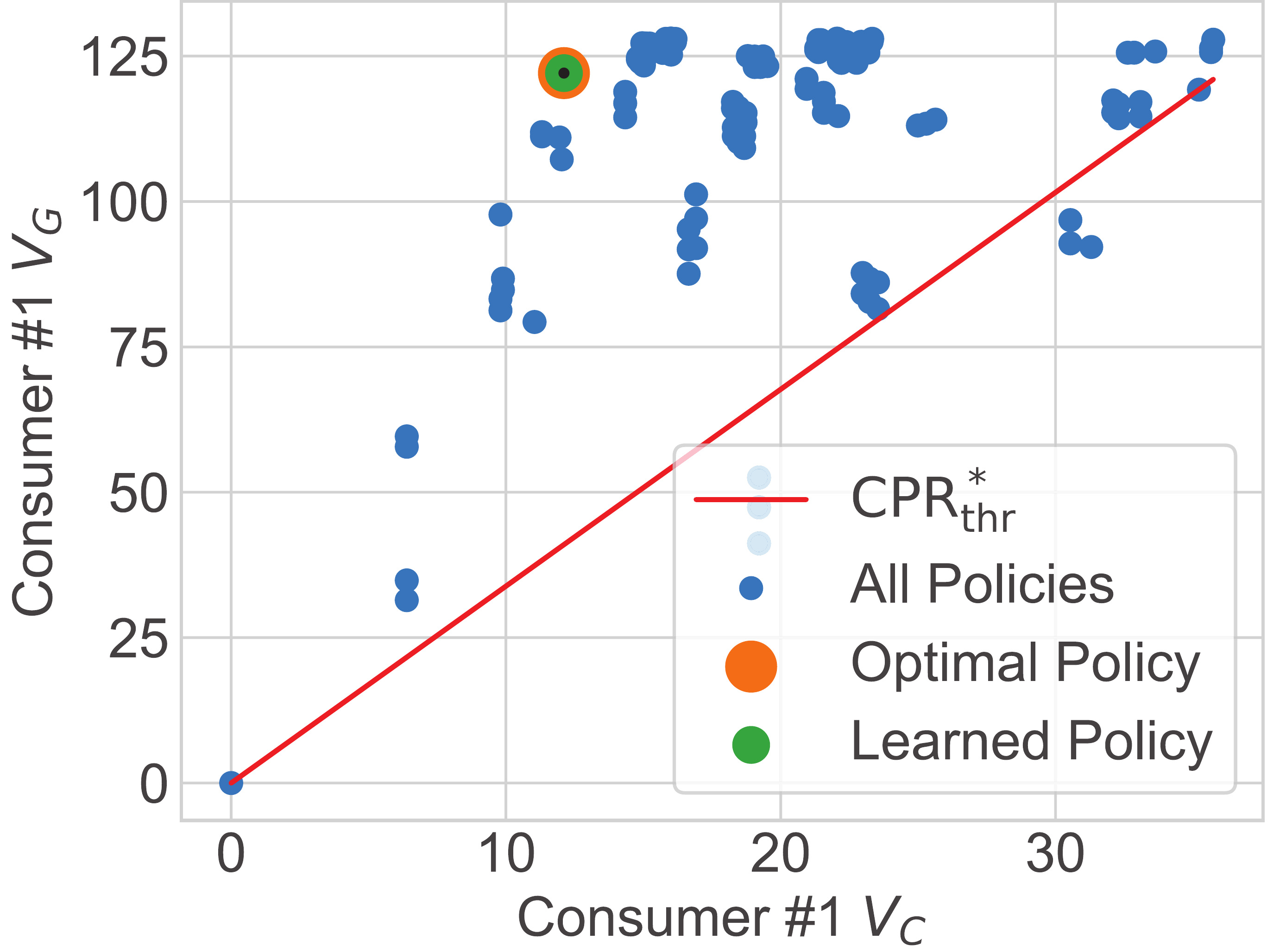}
}
\subfigure[user \#2's learned policy.] {
\label{Figure:case-2-policy}
\includegraphics[width=2.1in,angle=0]{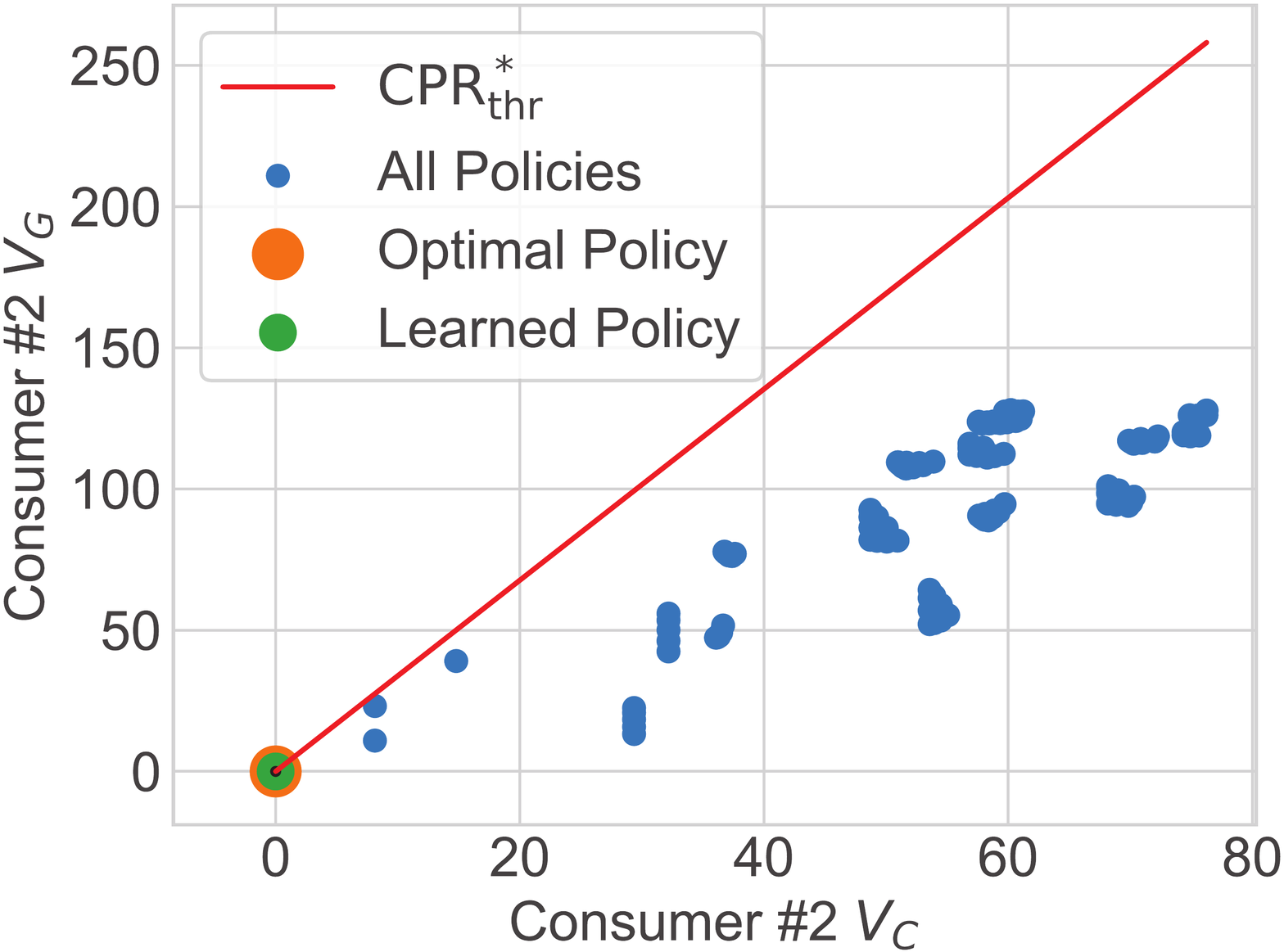}
}
\subfigure[user \#3's learned policy.] {
\label{Figure:case-3-policy}
\includegraphics[width=2.1in,angle=0]{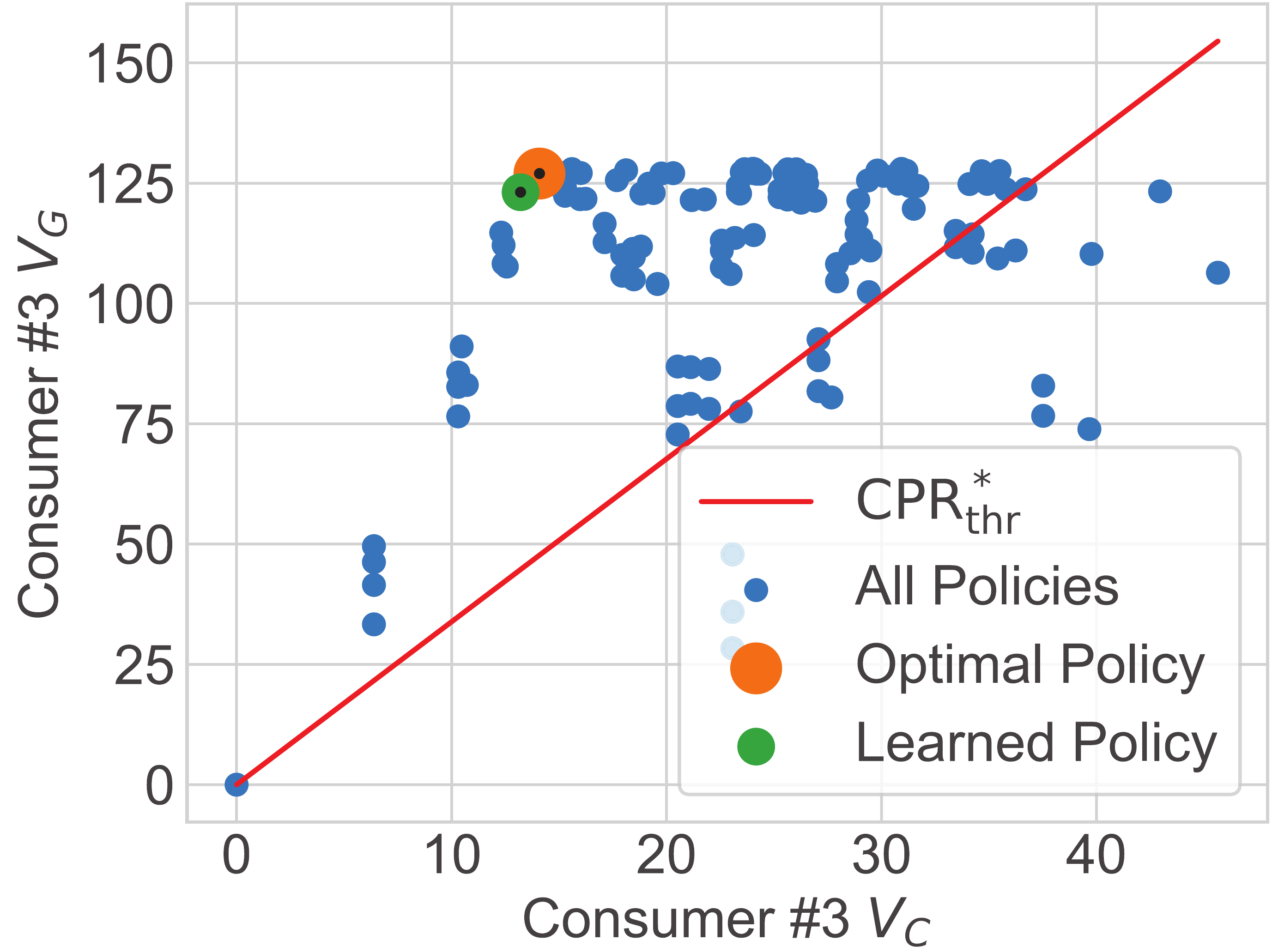}
}
\caption{Three examples of the convergence of each $\pi_i^*$ given a fixed $\text{CPR}_\text{thr}$.}
\label{Figure:user-policy}
\end{figure*}

As shown in Figure \ref{Figure:user-policy}, given a $\text{CPR}_\text{thr}$, the learned advertising policy $\pi$ of our \emph{MSBCB} converges to the optimal $\pi_j^*$. In Figure \ref{Figure:user-policy}, the x-axis denotes the cumulative cost, the y-axis denotes the cumulative value and the dots in blue represent the cumulative values and costs of all possible policies for each user. The red line represents $y=\text{CPR}_\text{thr}*x$, whose slope is $\text{CPR}_\text{thr}$. The orange point represents the optimal policy $\pi_i^*$ computed by enumerating all possible policies (blue points) and finding the one which maximize $V_{G}(i|\pi_i) - \text{CPR}_\text{thr}*V_{C}(i|\pi_i)$ according to \textbf{Theorem 1}. The green point denotes the learned policy of \emph{MSBCB}. \emph{In theory, the point of the optimal policy is the one whose $\text{CPR} > \text{CPR}^*_{thr}$ and vertical distance is the farthest from the red line.} A proof is provided in the \textbf{Theorem 4} in the later part. We present 3 convergence examples of different types in Figure \ref{Figure:user-policy}. In Figure \ref{Figure:user-policy} (a) and (b), the learned $\pi$ by the RL algorithm is exactly the same with the optimal $\pi^*$. In Figure \ref{Figure:user-policy} (b), the optimal policy is do not advertise to this user. In Figure \ref{Figure:user-policy} (c), the learned $\pi$ is approximately optimal.
Detail convergence statistics on the proportion of users whose policies converged to the optimal ones among all users are shown in Table \ref{tab:num of optimal solutions}.
For each user, we denote the vertical distance of the learned policy to the $\text{CPR}^*_{thr}$ line as $\text{dis}^*_{learned}$ and the vertical distance of the optimal policy $\pi^*$ to the $\text{CPR}^*_{thr}$ line as $\text{dis}^*_{optimal}$. We denote $\text{R}^*_{opt}=\text{dis}^*_{learned} / \text{dis}^*_{optimal}$ as the approximation ratio. According to \textbf{Theorem 4}, if the $\text{R}^*_{opt}$ is 100\%, then the learned strategy is exactly the optimal strategy. Otherwise, we denote that the learned strategy is the $\text{R}^*_{opt}$-approximation strategy. As shown in Table \ref{tab:num of optimal solutions}, there are 74.9\% policies achieve more than 90\%-approximation ratios and 53.3\% policies achieve exactly the optimal.
\begin{table*}[htbp]
  \centering
  \caption{Optimal types of each $\pi_i^*$ of 10000 users}
  \label{tab:num of optimal solutions}
  \resizebox{0.5\columnwidth}{!}{
  \begin{tabular}{lccc}
    \toprule

    $\text{R}^*_{opt}$ & 100\% & [90\%, 100\%)   & [0\%, 90\%) \\
    \midrule
    Percentage & 53.3\%  & 21.6\%   &   25.3\%      \\
    \bottomrule
  \end{tabular}}
\end{table*}

\textbf{Theorem 4.} \emph{The point of the optimal policy is the one whose $\text{CPR} > \text{CPR}^*_{thr}$ and vertical distance to $\text{CPR}^*_{thr}$ line (red line) is the farthest among all policy dots in Figure \ref{Figure:user-policy}.}

\begin{figure}[htbp]
  \centering
  \includegraphics[scale=0.5]{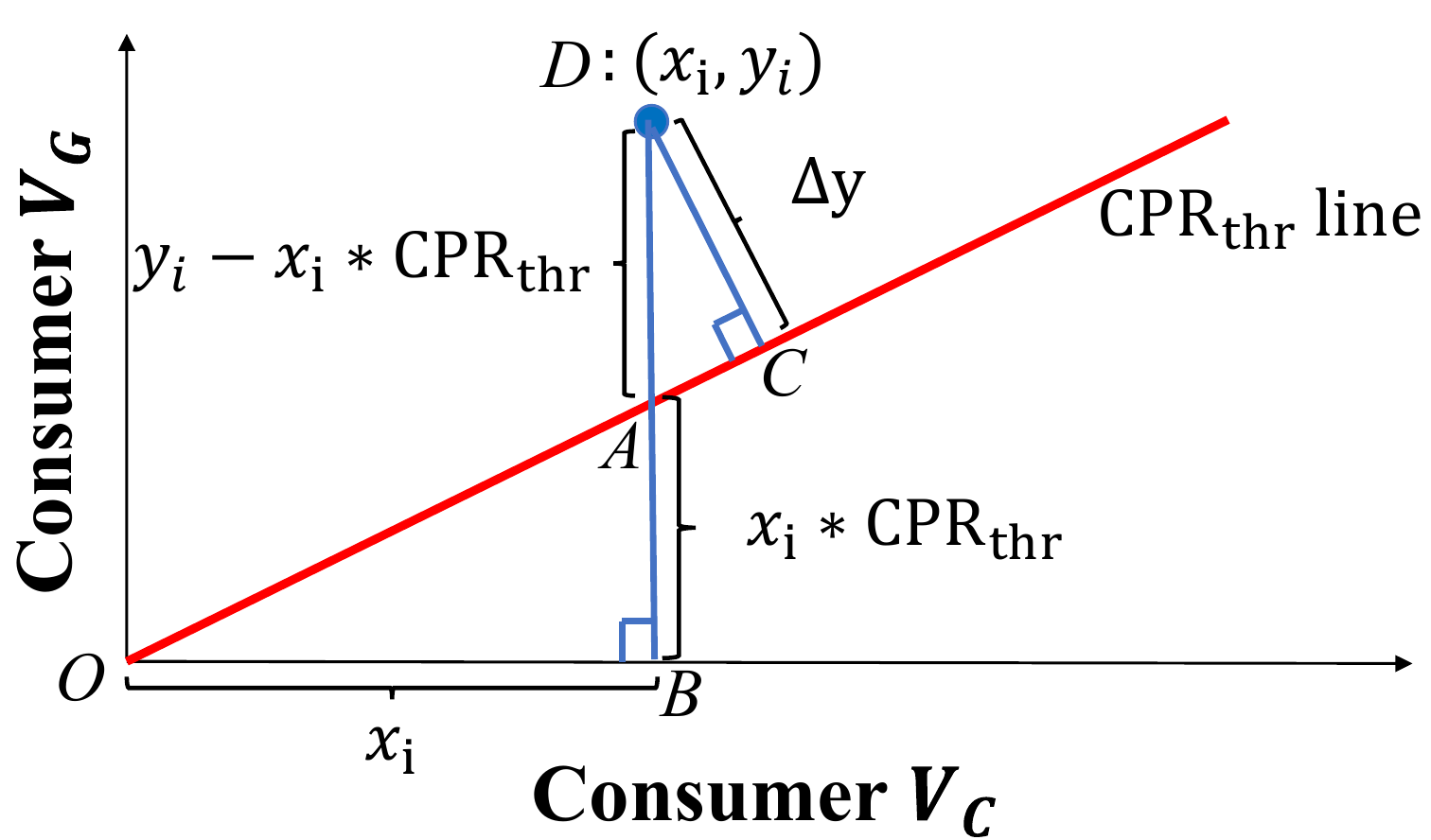}
  \caption{Proof of Optimal Policy Dot}
  \label{Proof Optimal Policy Dot}
\end{figure}

\emph{Proof.} Here we give a simple proof of \textbf{Theorem 4}.
As we can see in Figure \ref{Proof Optimal Policy Dot}, the blue dot $D:(x_{i}, y_{i})$ is an arbitrary policy $i$ in Figure \ref{Figure:user-policy}. Suppose the vertical distance of $D$ to $CPR^*_{thr}$ line (red line) is $\Delta y_{i}$ (segment $DC$ in the figure). We then draw a vertical line of x-axis from dot $D$ to dot $B$. We can then calculate the length of segments: $OB=x_{i}$, $BA=x_{i}*\text{CPR}_{thr}$, $DA=y_{i}-x_{i}*\text{CPR}_{thr}$. It's evident that $\bigtriangleup OAB \sim \bigtriangleup DAC$, which means $\frac{DC}{OB}=\frac{DA}{OA}=\frac{DA}{\sqrt{(OB)^2+(AB)^2}}$.
We can derive that
\begin{equation}
    \frac{\Delta y_{i}}{x_{i}}=\frac{y_{i}-x_{i}*\text{CPR}_{thr}}{\sqrt{x^2_{i}+(x_{i}*\text{CPR}_{thr})^2)}}
\end{equation}

As $x_{i} > 0$, we can further derive that
\begin{equation}
\label{similar triangles theorem}
    \Delta y_{i} = \frac{y_{i}-x_{i}*\text{CPR}_{thr}}{\sqrt{1+\text{CPR}^2_{thr}}}
\end{equation}

Suppose the dot of a policy is $(x^{*}, y^{*})$, which has farthest vertical distance $\Delta y^{*}$ from the $CPR^*_{thr}$ line, that is, for a dot of arbitrary policy $i$, we have $\Delta y_{i} \leq \Delta y^{*}$.  According to Equation \ref{similar triangles theorem}, we have
\begin{equation}
    \frac{y_{i}-x_{i}*\text{CPR}_{thr}}{\sqrt{1+\text{CPR}^2_{thr}}} \leq \frac{y^{*}-x^{*}*\text{CPR}_{thr}}{\sqrt{1+\text{CPR}^2_{thr}}}
\end{equation}
Then we get $y_{i}-x_{i}*\text{CPR}_{thr} \leq y^{*}-x^{*}*\text{CPR}_{thr}$, which means $(x^{*}, y^{*})$ is the dot of the optimal policy. Thus, we complete the proof.

\subsubsection{Convergence of $\text{CPR}_\text{thr}^*$.} 
\begin{figure}[ht]
  \centering
  \includegraphics[width=2.8in]{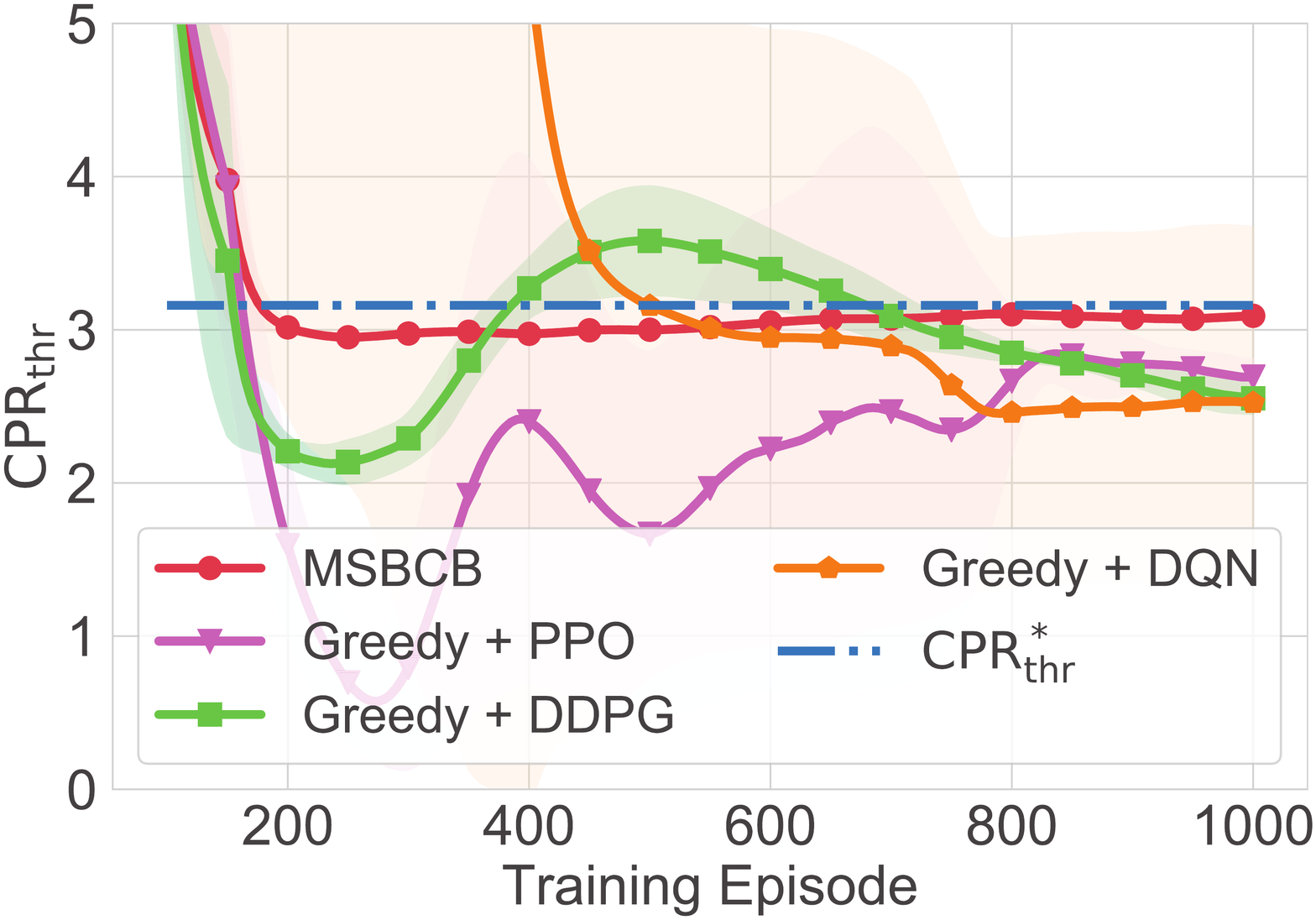}
  \caption{The convergence of $\text{CPR}_\text{thr}^*$}
  \label{convergence of CPR_thr}
\end{figure}

In Figure \ref{convergence of CPR_thr}, we plot the learning curves of the $\text{CPR}_\text{thr}$ of our \emph{MSBCB} as well as 3 RL approaches. The dotted blue line denotes the optimal $\text{CPR}_\text{thr}^*$ computed by the \emph{MSBCB (enum)} of Table 1 of paper. Figure \ref{convergence of CPR_thr} shows that the learned $\text{CPR}_\text{thr}$ of our \emph{MSBCB} could gradually converge to the optimal $\text{CPR}_\text{thr}^*$ approximately, which is much better than the other 3 RL approaches.

\subsection{Gap to Market Second Price.}
\begin{figure}[htbp]
  \centering
  \includegraphics[width=2.8in]{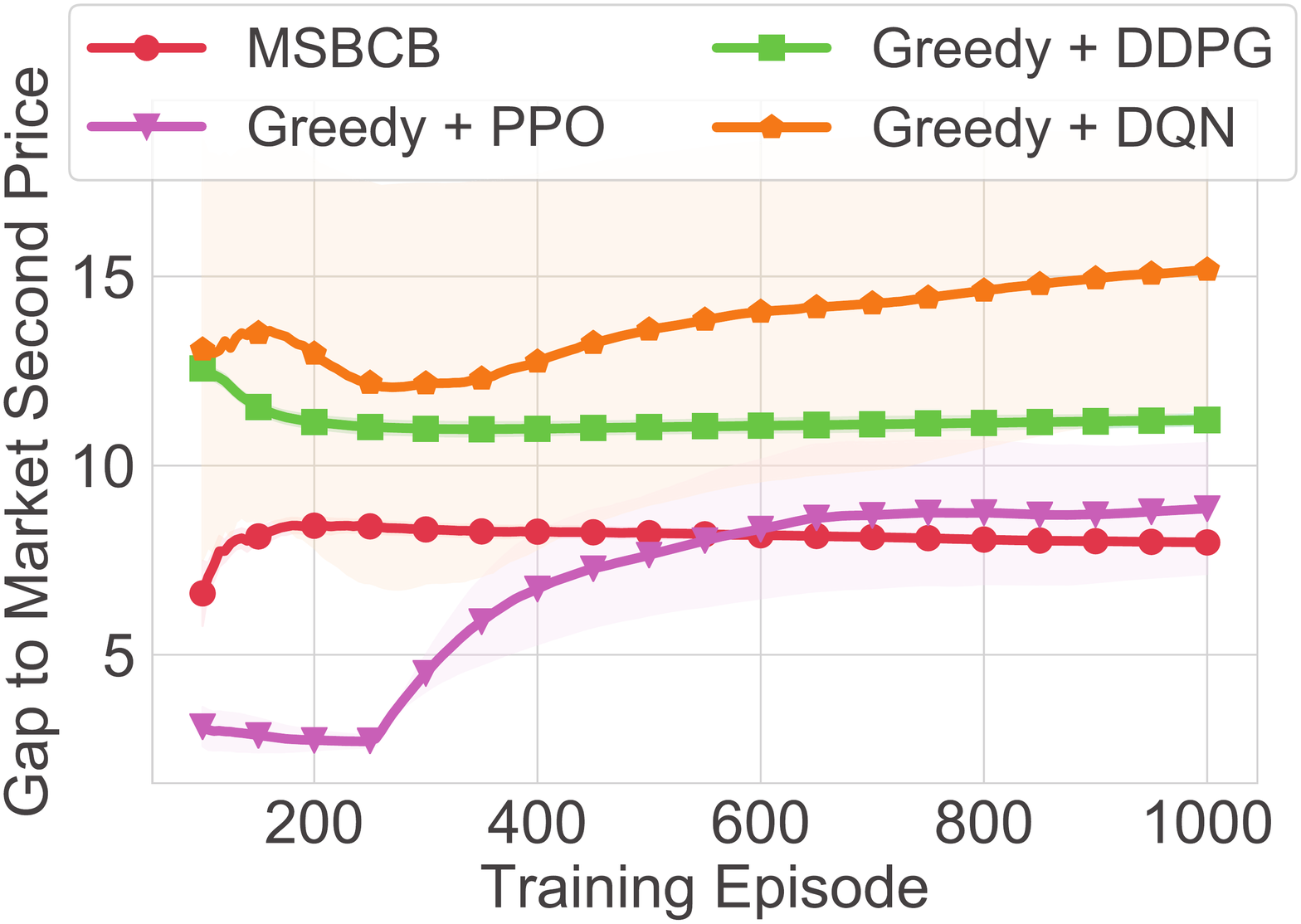}
  \caption{The average gaps of the bids to the second prices in the auction by different methods.}
  \label{Gap To Second Price}
\end{figure}

 Figure \ref{Gap To Second Price} shows average gaps between the bid of the agent of different approaches and the second price in the auction. Results indicate that the bid prices given by the \emph{MSBCB} agent are closer to the second price in the auction, which can reduce the risk of economic loss when the market price fluctuates.

\subsection{Effectiveness of Action Space Reduction.}
\begin{figure}[htb]
  \centering
  \includegraphics[width=3.2in]{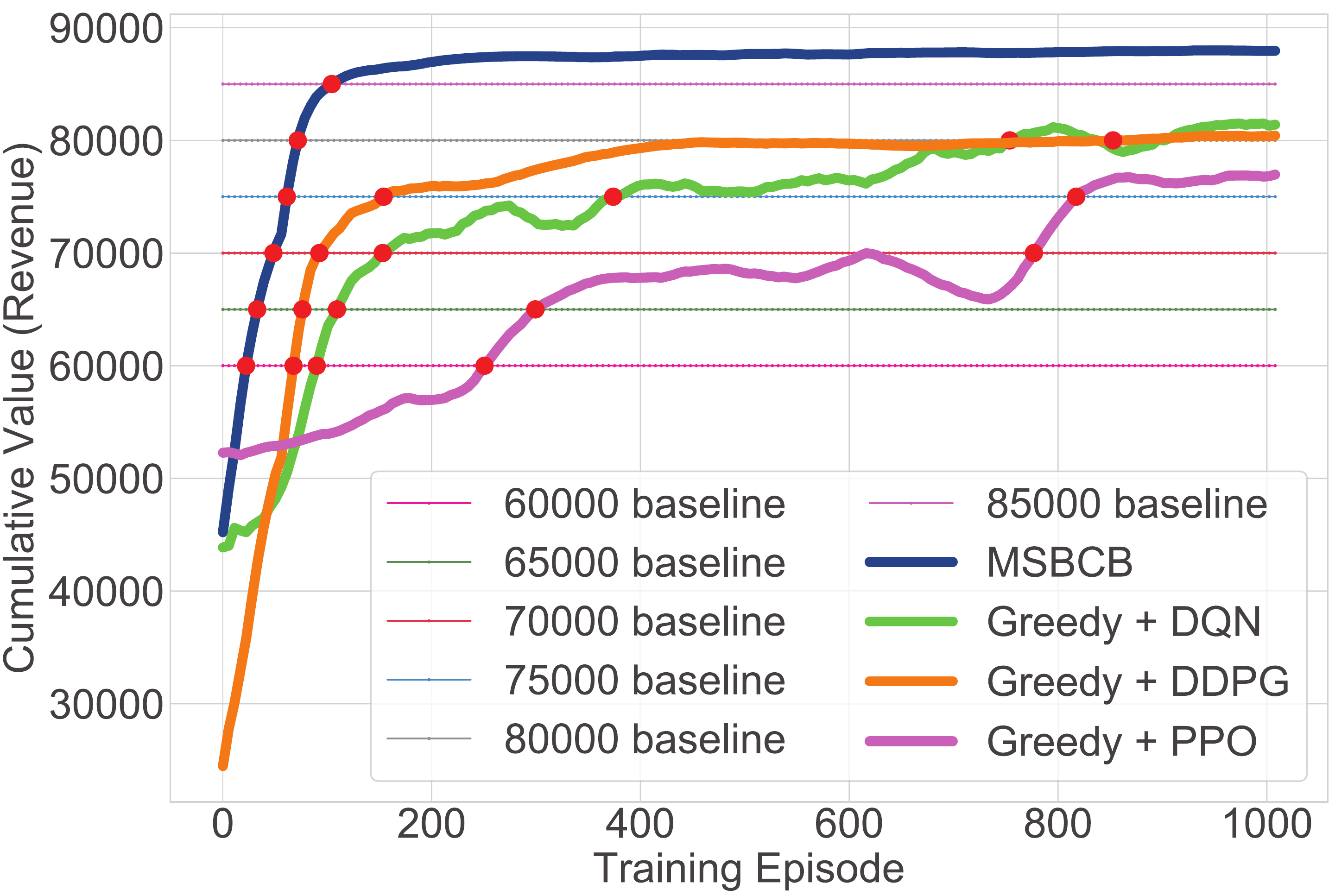}
  \caption{The comparison of the number of training episodes needed by different approaches when achieving the same revenue level.}
  \label{figure:sample utilization}
\end{figure}

Here we give a more detailed comparison of \emph{MSBCB} and RL baselines to demonstrate the effectiveness of action space reduction. As shown in Figure \ref{figure:sample utilization} and Table \ref{tab:sample utilization}, \emph{MSBCB} (with action space reduction) can reach exactly the same cumulative value much more quickly than the other 3 RL baselines. \emph{MSBCB} can reach a cumulative value of 85000 in only 104 epochs, which proves that action space reduction can effectively improve the sample utilization to converge to higher performance with faster speed.
\begin{table*}[htbp]
  \caption{The training epochs and the number of samples needed by different approaches when achieving the same revenue level.}
  \label{tab:sample utilization}
  \scalebox{0.75}{
  \begin{tabular}{l|cc|cc|cc|cc|cc|cc}
    \toprule

    Cumulative Value & \multicolumn{2}{|c}{60000} & \multicolumn{2}{|c}{65000} & \multicolumn{2}{|c}{70000} & \multicolumn{2}{|c}{75000} & \multicolumn{2}{|c}{80000} & \multicolumn{2}{|c}{85000} \\
    Method     & \#Epoch & \#Samples  & \#Epoch & \#Samples & \#Epoch & \#Samples & \#Epoch & \#Samples & \#Epoch & \#Samples & \#Epoch & \#Samples \\
    \midrule
    Greedy + PPO        &  251 & 1280000   &  299 & 1530880   &  776 & 3973120           & 817 & 4183040 & - & - & - & -        \\
    Greedy + DDPG       &  68 & 343040     &  76 & 389120    &  92 & 471040    & 154 & 788480  &  853 & 4362240  & - & -      \\
    Greedy + DQN        &  90 & 455680     &  109 & 558080   &  153 & 783360   & 373 & 1909760  & 754 & 3855360  & - &-          \\
    \textbf{MSBCB}      &  \textbf{22} & \textbf{112640}     &  \textbf{33} & \textbf{163840}    &  \textbf{48} & \textbf{245760}    & \textbf{61} & \textbf{312320}   & \textbf{71} & \textbf{363520}    & \textbf{104} & \textbf{532480}        \\
    \bottomrule
  \end{tabular}}
\end{table*}

\section{Empirical Evaluation: Supplementary of Online A/B Testing}
\label{online_ab_testing}
In online A/B Testing, we conduct further analyses to verify the effectiveness of our \emph{MSBCB} and find out whether our approach could benefit most advertisers.

Firstly, we analyze the performance of our \emph{MSBCB} for each advertiser. To guarantee the statistical significance, only the advertisers with more than 100 conversions in a week are included.
The detail results of top-10 advertisers with the largest costs are shown in Table \ref{tab:shop_results}. In Table \ref{tab:shop_results}, under the same budget constraint, our \emph{MSBCB} can increase the Revenues and ROIs of most advertisers compared with the myopic \emph{Contextual Bandit} approach. Although the ROI of advertiser 7 drops slightly, our \emph{MSBCB} contributes to much more PVs (Page Views).

\begin{table}[htbp]
   \centering
   \resizebox{0.5\columnwidth}{!}{%
   \begin{tabular}{c | c c  c  c  c }
      \toprule
            & Revenue & Cost & CVR & PV & ROI \\
      \midrule
      Advertiser 1       & 5.1\%  &-6.3\%  & 17.2\% & 9.6\%  & 12.2\%\\
      Advertiser 2       & 7.5\%  & 2.1\%  & 5.2\%& 12.2\%  & 5.3\%\\
      Advertiser 3       & 48.6\%  & 10.9\%  & 27.6\% & 28.9\%  & 33.9\%\\
      Advertiser 4       & 3.1\%  & 2.8\%  & 1.1\% & 9.6\%  & 0.3\%\\
      Advertiser 5       & 12.7\%  & 1.7\%  & 12.9\% & 17.8\% & 10.8\%\\
      Advertiser 6       & 10.8\%  & 2.2\% & 4.4\% & 13.8\%  & 8.4\%\\
      Advertiser 7       & 1.9\%  & 3.8\%  & 4.6\% & 31.5\%  & -1.8\%\\
      Advertiser 8       & 5.6\%  &-4.8\%  & 2.9\% & 10.7\% & 11.1\%\\
      Advertiser 9       & 6.7\%  & -2.4\%  &  6.3\% & 21.0\%  & 9.4\%\\
      Advertiser 10       & 5.8\%  & -0.8\%  &  2.5\% & 8.0\%  & 6.7\%\\
      \bottomrule
   \end{tabular}%
   }
   \caption{The improvements in Revenue, CVR, PV and ROI of our \emph{MSBCB} compared with the myopic \emph{Contextual Bandit} method.}
   \label{tab:shop_results}
\end{table}

\begin{figure}[htb]
   \centering
   \includegraphics[width=0.4\columnwidth]{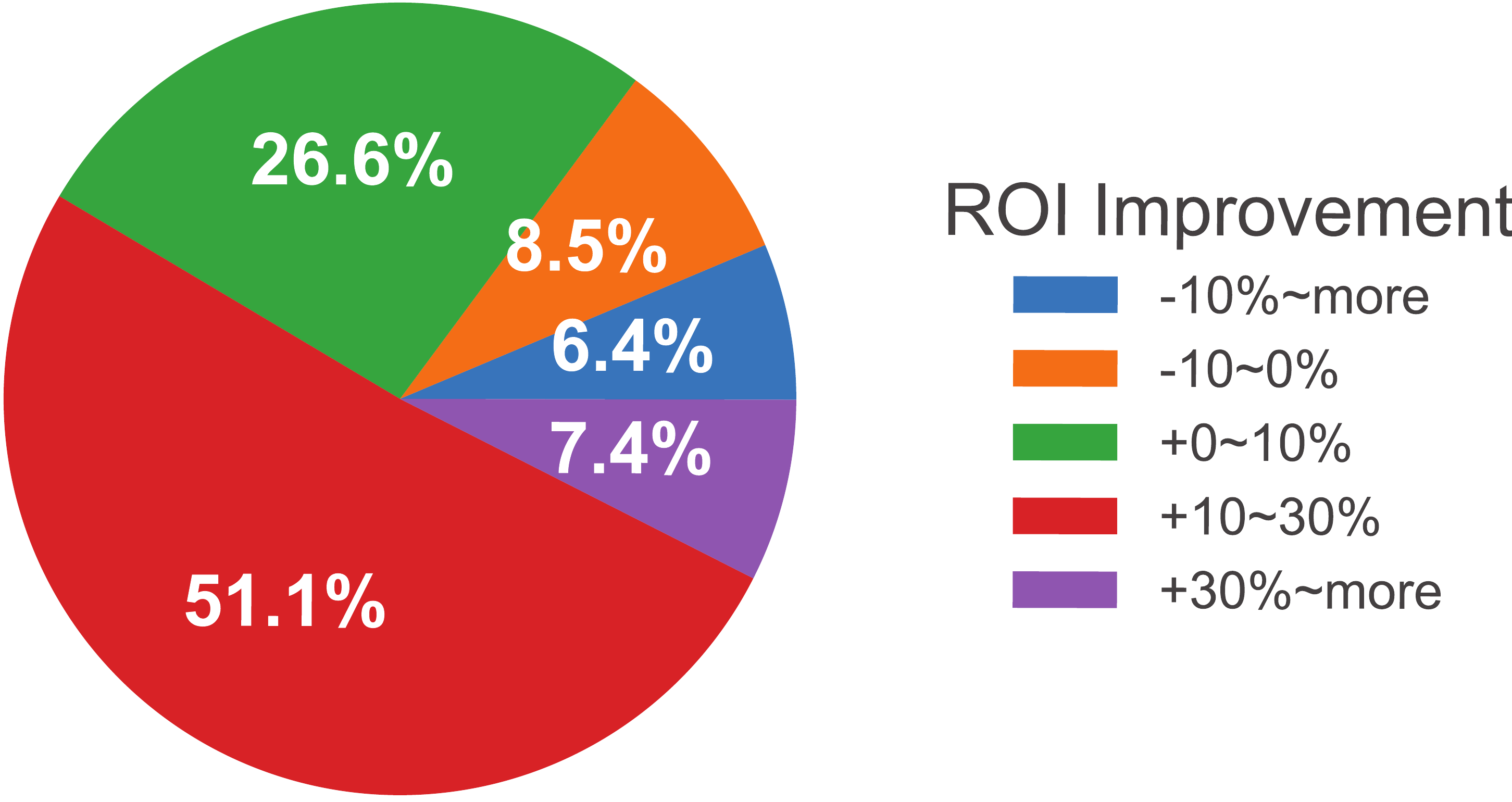}
   \caption{The distribution of ROI improvements for all advertisers of our \emph{MSBCB} compared with the myopic \emph{Contextual Bandit} method.}
   \label{fig:roi_details}
\end{figure}

Besides, in Figure \ref{fig:roi_details}, we give the detail proportions of advertisers whose ROIs are improved. Among all advertisers, 85.1\% advertisers obtain positive ROI improvements while the rest of 14.9\% advertisers are in the so-called quantity and quality exchange situations: their PV increments are larger than the ROI drops. We say that it’s also acceptable for some advertisers because the PV increments might lead to secondary exposures to an advertiser and thus lower the ROI within the current time period. But the increase in PV may leave deeper impressions to the users and contribute to the long-term future revenues. In addition, Figure \ref{fig:roi_details} demonstrates that our \emph{MSBCB} can be well applied to the multi-agent setting (which involves multiple advertisers) in the real-world auction environment, which could increase the overall revenue for most advertisers.

\begin{figure}[htb]
   \centering
   \includegraphics[width=0.5\columnwidth]{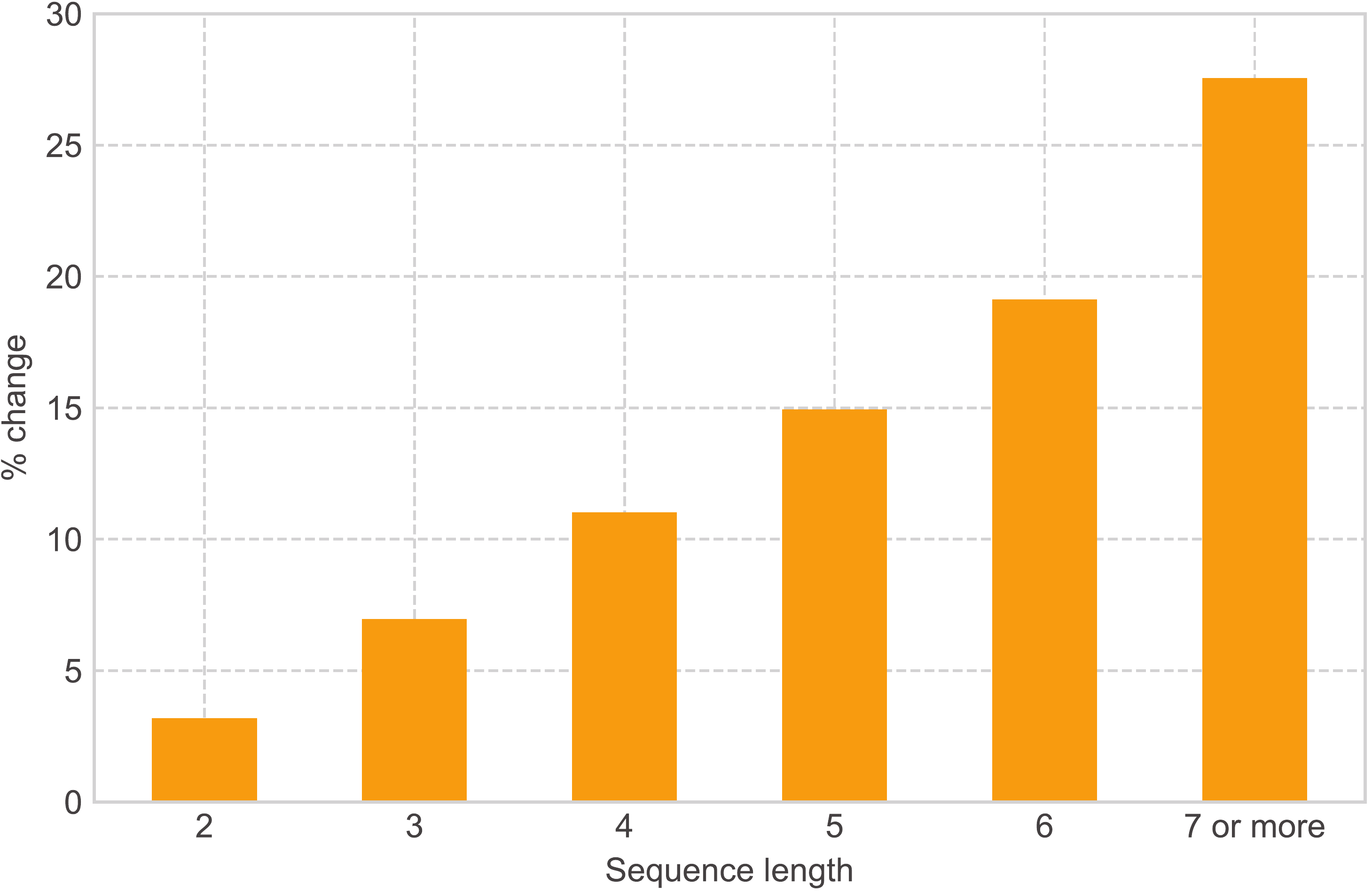}
   \caption{The proportion improvements in the sequence length of our \emph{MSBCB} compared with the myopic \emph{Contextual Bandit} method.}
   \label{fig:seq_len}
\end{figure}

In order to highlight the advantage of our method in long-term revenue optimization, we compared the average number of times (we call the sequence length) that a user contact with an advertisement under different approaches. Figure \ref{fig:seq_len} shows the extent of \emph{MSBCB}'s improvement relative to \emph{Contextual Bandit} in the proportion of the user sequence length. The results show that our \emph{MSBCB} can increase the proportion of the sequences with larger sequence length. Especially, the ratio of sequence length of 7 is increased by nearly 30\%. It shows that our method can promote longer user behavior sequences, and longer user behavior sequence means more opportunities to affect the user's mentality towards an advertisement, thereby improving the long-term revenue for an advertisement.

\begin{figure*}[h!]
   \centering
   \includegraphics[width=0.9\columnwidth]{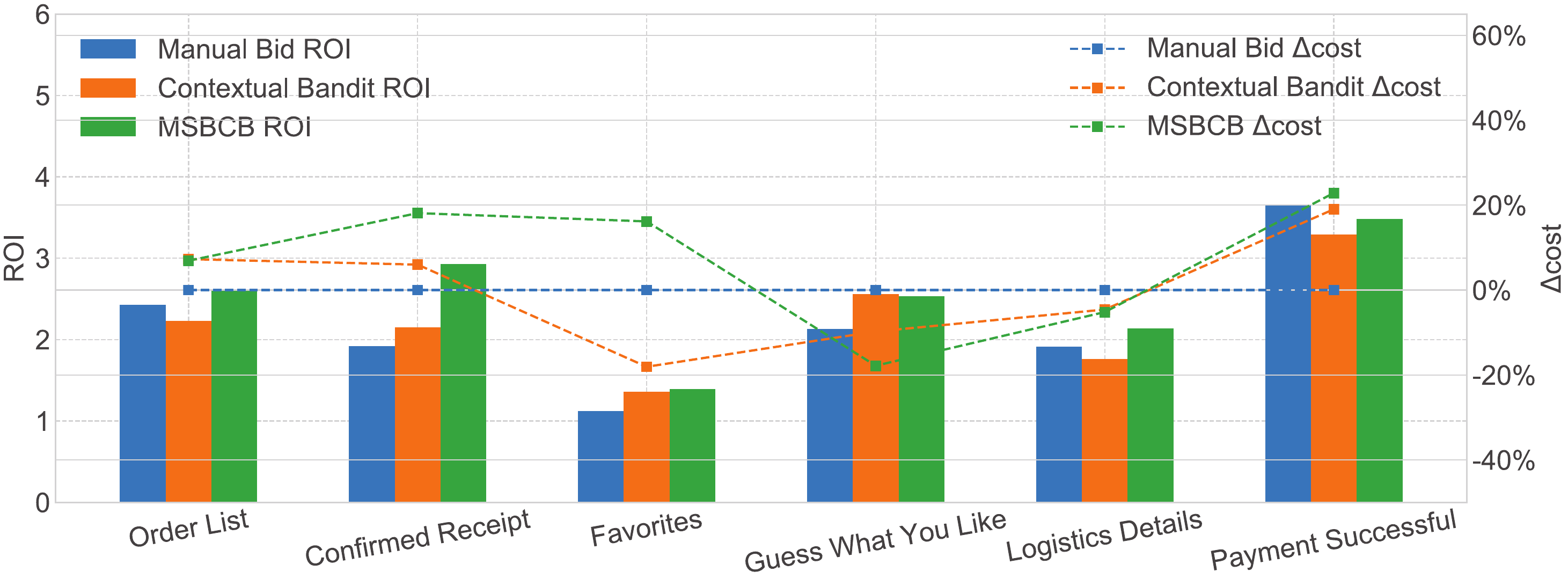}
   \caption{ROI and budget allocation among different channels. }
   \label{fig:roi_percent}
\end{figure*}

Further, we also analyze the ROI performances of the compared 3 algorithms (i.e., \emph{CEM}, \emph{Contextual Bandit} and our \emph{MSBCB}) in different channels. Figure \ref{fig:roi_percent} shows the budget allocation distributions of all approaches among 6 channels and the corresponding ROIs. The left axis represents the ROI, and the ROI performances of each algorithm among different channels are given by the corresponding bar charts. The right axis represents the increments or decrements of the actual costs of \emph{MSBCB} and \emph{Contextual Bandit} relative to \emph{CEM}, which are indicated by the line charts.
In Figure \ref{fig:roi_percent}, we observe the following two phenomena:
\begin{enumerate}[1)]
\item \emph{MSBCB} and \emph{Contextual Bandit} both spend more budgets on channels with higher ROIs, especially on the \emph{Payment Successful} channel, where the average ROI is much higher.
\item Compared with \emph{Contextual Bandit}, \emph{MSBCB} allocates more budget from the \emph{Guess What You Like} channel to other channels, especially the \emph{Favorites} channel, \emph{Confirmed Receipt} channel and the \emph{Payment Successful} channel.
\end{enumerate}
These phenomena show that our \emph{MSBCB} can reasonably allocate budgets among different channels and spend more budgets in channels with higher ROIs. In addition, compared with the myopic method \emph{Contextual Bandit}, our long-term \emph{MSBCB} is more optimistic about channels during and after purchasing, which shows that our \emph{MSBCB} prefers a longer interaction sequence to optimize cumulative long-term values.
\end{appendices}
\end{document}